%% file: main_file.tex
\documentclass[sigplan,10pt,twocolumn]{acmart}

\acmSubmissionID{125}
\acmYear{2026}\copyrightyear{2026}
\setcopyright{cc}
\setcctype[4.0]{by}
\acmConference[EUROSYS '26]{European Conference on Computer Systems}{April 27--30, 2026}{Edinburgh, Scotland Uk}
\acmBooktitle{European Conference on Computer Systems (EUROSYS '26), April 27--30, 2026, Edinburgh, Scotland Uk}
\acmDOI{10.1145/3767295.3769328}
\acmISBN{979-8-4007-2212-7/26/04}
\input{others/packages}

\input{others/configs}
\begin{document}

%don't want date printed
\date{}

\title[\method: Responsive LLM Text Streaming Serving under Request Burst \\via Preemptive Scheduling]{\method: Responsive LLM Text Streaming Serving under Request Burst via Preemptive Scheduling}

\input{others/authors}
\input{text/0_abstract}




\maketitle

\renewcommand{\shortauthors}{Junyi Chen et al.}
\newcommand{\mypara}[1]{\vspace{0.05cm}\noindent{\bf {#1}.}~}
\newcommand{\myparacolon}[1]{\vspace{0.05cm}\noindent{\bf {#1}:}~}

\input{text/1_intro}
\input{text/2_background}

\input{text/3_overview}

\input{text/4_scheduler}

\input{text/5_memory_management}
\input{text/6_implementation}
\input{text/7_experiment}
\input{text/discussion}
\input{text/8_related}
\input{text/9_conclusion}

\begin{acks}
This work was sponsored in part by the National Key R\&D Program of China (No. 2022ZD0119100), in part by China NSF grant No. 62472278, 62025204, 62432007, 62441236, 62332014, and 62332013, and in part by Tencent Rhino Bird Key Research Project. 
This work was partially supported by SJTU Kunpeng \& Ascend Center of Excellence.
The authors would like to thank SenseTime for providing computational resources for this work.
The opinions, findings, conclusions, and recommendations in this paper are those of the authors and do not necessarily reflect the views of the funding agencies or the government.
\end{acks}

\newpage
\bibliographystyle{plain}
\balance
\bibliography{reference}

\end{document}

%% file: others/packages.tex
\usepackage{titlesec}
\usepackage{mathrsfs}
\usepackage{amsmath,amsfonts,amsthm}
\usepackage{enumitem} 
\usepackage{graphicx}
\usepackage{extarrows}

\usepackage{amssymb}
\usepackage{subfigure}
\usepackage{verbatim}
\usepackage{makecell}
\usepackage{diagbox}
\usepackage{bm}
\usepackage{multirow}
\usepackage{float}
\usepackage{placeins}
\usepackage[linesnumbered, ruled]{algorithm2e}
\SetAlFnt{\small}
\usepackage{epstopdf}
\usepackage{setspace}
\usepackage[super]{nth}
\usepackage{fancyhdr}
\usepackage{slashbox}
\usepackage{lipsum,multicol}
\usepackage{url}
\usepackage{xcolor}
\usepackage{booktabs}
\usepackage{diagbox}
\usepackage[skip=5pt]{caption}
\usepackage{ragged2e}
\usepackage{wrapfig}
\usepackage{tikz}
\usepackage{hyperref}
\usepackage{graphicx}
\usepackage[normalem]{ulem}

%% file: others/configs.tex
\def \eg {\emph{e.g.}, }
\def \ie {\emph{i.e.}, }

\SetCommentSty{mycommfont}

\newcommand*\blackcircled[1]{\tikz[baseline=(char.base)]{
            \node[shape=circle,fill,inner sep=1pt] (char) {\textcolor{white}{#1}};}}

\newcommand{\method}{{TokenFlow}\xspace}

%% file: others/authors.tex
{%for single author (just remove % characters)
\author{Junyi Chen}
\orcid{0009-0003-7397-6311}
\affiliation{
\institution{Shanghai Jiao Tong University}
\city{Shanghai}
\country{China}}
\email{junyi.chen@sjtu.edu.cn}

\author{Chuheng Du}
\orcid{0009-0005-3465-4023}
\affiliation{
\institution{Shanghai Jiao Tong University}
\city{Shanghai}
\country{China}}
\email{dch7723@sjtu.edu.cn}

\author{Renyuan Liu}
\orcid{0000-0001-9710-6116}
\affiliation{
\institution{George Mason University}
\city{Fairfax, VA}
\country{USA}}
\email{rliu23@gmu.edu}

\author{Shuochao Yao}
\orcid{0000-0001-9710-6116}
\affiliation{
\institution{George Mason University}
\city{Fairfax, VA}
\country{USA}}
\email{shuochao@gmu.edu}

\author{Dingtian Yan}
\orcid{0009-0004-3112-8215}
\affiliation{
\institution{China Telecom Corporation Limited Shanghai Branch}
\city{Shanghai}
\country{China}}
\email{yandt@chinatelecom.cn}

\author{Jiang Liao}
\orcid{0009-0007-2128-3096}
\affiliation{
\institution{China Telecom Corporation Limited Shanghai Branch}
\city{Shanghai}
\country{China}}
\email{liaojiang.sh@chinatelecom.cn}

\author{Shengzhong Liu}
\authornote{Shengzhong Liu is the corresponding author.}
\orcid{0000-0002-7643-7239}
\affiliation{
\institution{Shanghai Jiao Tong University}
\city{Shanghai}
\country{China}}
\email{shengzhong@sjtu.edu.cn}

\author{Fan Wu}
\orcid{0000-0003-0965-9058}
\affiliation{
\institution{Shanghai Jiao Tong University}
\city{Shanghai}
\country{China}}
\email{fwu@cs.sjtu.edu.cn}

\author{Guihai Chen}
\orcid{0000-0002-6934-1685}
\affiliation{
\institution{Shanghai Jiao Tong University}
\city{Shanghai}
\country{China}}
\email{gchen@cs.sjtu.edu.cn}
} % end author

%% file: text/0_abstract.tex
\begin{abstract}
%This paper proposes a
% Recent advances in large language models (LLMs) have enabled a wide range of real-time interactive applications, including AI assistants, customer service agents, and voice-based assistants. 
% time-to-first-token (TTFT) for responsiveness and steady per-token generation rate for stalling avoidance.

Real-time LLM interactions demand \textit{streamed token generations}, where text tokens are progressively generated and delivered to users while balancing two objectives: 
responsiveness (\ie low time-to-first-token) and steady generation (\ie required time-between-tokens).
Standard LLM serving systems suffer from the inflexibility caused by non-preemptive request scheduling and reactive memory management, leading to poor resource utilization and low request processing parallelism under request bursts.
Therefore, we present \method, a novel LLM serving system with enhanced text streaming performance via preemptive request scheduling and proactive key-value (KV) cache management. 
\method dynamically prioritizes requests based on real-time token buffer occupancy and token consumption rate, while actively transferring KV cache between GPU and CPU memory in the background and overlapping I/O with computation to minimize request preemption overhead. 
Extensive experiments on Llama3-8B and Qwen2.5-32B across multiple GPUs (RTX 4090, A6000, H200) demonstrate that \method achieves up to 82.5\% higher effective throughput (accounting for actual user consumption) while reducing P99 TTFT by up to 80.2\%, without degrading overall token throughput.
\end{abstract}

% \keywords{Do, Not, Us, This, Code, Put, the, Correct, Terms, for,

%% file: text/1_intro.tex
\section{Introduction}\label{sec:intro}
% background
Recent Large Language Models (LLMs) such as GPT~\cite{radford2019language,brown2020language,achiam2023gpt}, LLaMA~\cite{touvron2023llama,grattafiori2024llama}, Qwen~\cite{bai2023qwen,yang2024qwen2}, and DeepSeek~\cite{liu2024deepseek,guo2025deepseek} have demonstrated remarkable capabilities across a wide range of language processing tasks, which have led to a surge in LLM-powered applications, especially those requiring real-time interactions. 
Examples include AI assistants~\cite{sergeyuk2025using,schon2023ai}, intelligent customer service agents for E-commerce~\cite{10.1145/3703187.3703197} and finance~\cite{10.1145/3688399}, voice-driven assistants~\cite{wang2024task}, external tool invocation~\cite{chen-etal-2025-pre3}, and collaborative productivity tools~\cite{10.1145/3663384.3663398,xu2025novarealtimeagenticvisionlanguage}. 
As these applications shift from offline batch processing to online real-time interactions, the demand for low-latency and high-throughput LLM serving is growing rapidly. 

% Context and Problem Setup 
Unlike traditional machine learning tasks that generate output in a single forward pass, LLM generates and delivers tokens progressively to users, also called \textit{text streaming}, as the process is analogous to video streaming in content delivery networks: a short initial delay (time-to-first-token) is acceptable, but subsequent tokens should be generated fast enough to match user consumption speed. 
An \textit{output buffer} stores generated but not yet consumed text tokens. {This is where we find our key optimization opportunity: because an LLM's generation speed is typically much faster than a user's reading speed, there will inevitably be a surplus of tokens. These extra tokens can be efficiently stored in the output buffer, providing a valuable cushion. By leveraging this buffer, our system gains more opportunities to optimize efficiency and performance. However, this buffer-based approach also presents a critical challenge:}
If the output buffer becomes empty before new tokens are ready, users experience stalling or visible latency spikes, disrupting the interaction flow. 
Conversely, aggressive token generations that ignore actual user consumption rates produce unbalanced responses among requests upon request burst: Actively serving requests receive token generations at a rate exceeding user comprehension, while queuing requests experience high TTFT before receiving their first response. 
% We argue that LLM serving performance hinges on not only the model inference speed but also the matching level between token generation and user consumption rates.
Orthogonal to model acceleration, this paper works on maximizing the request process parallelism and optimizing the service responsiveness upon request burst via matching the request token generation rate with its respective user consumption rate.

% Challenges and limitations of current approaches
The core challenge in LLM streaming systems~\cite{liu2024andes,10.1145/3662006.3662063,xiao2025streaming,10.1145/3620665.3640383,yao2025deltazip} lies in balancing the two competing objectives: minimizing \textit{time-to-first-token (TTFT)} for service responsiveness while sustaining low \textit{time-between-tokens (TBT)} for smooth user comprehension. 
Standard LLM serving systems fail to optimize both objectives during request bursts: The rigid first-come-first-served (FCSF) scheduling in SGLang~\cite{zheng2024sglang} causes unacceptable queueing delays.
% while the responsiveness improvements by QoS-aware systems like Andes~\cite{} come at the cost of degraded resource utilization as their reactive memory management introduces non-ignorable I/O delays. 
Although preemptive scheduling between request reduces queueing delays, its direct application (as in Andes~\cite{liu2024andes}) induces frequent request context switches\footnote{Context switch here means move the KV cache of evicted requests out of GPU memory and move in the KV cache of selected requests.} that interfere with LLM decoding computation and achieve limited resource utilization.
Meanwhile, as more requests are served alternatively on the GPU, their KV cache storage can easily overload the limited GPU memory and turn into a memory-bounded problem.
Therefore, a novel LLM serving system integrating flexible preemptive request scheduling with effective memory management supporting seamless request preemption-resumption cycles is needed.

% These limitations ultimately stem from three fundamental gaps: scheduling inflexibility that ignores dynamic consumption rates, memory management incapable of seamless preemption-resumption cycles, and evaluation metrics that fail to properly capture streaming quality degradation.
% Achieving this balance requires careful coordination between request scheduling and memory management subsystems. 
% On the scheduling front, effective load balancing must dynamically account for both token buffer occupancy and output speed constraints, which critically determine the trade-off between initial responsiveness and streaming stability. 
% For memory management, the system should minimize I/O overhead by strategically overlapping data transfers with computation to ensure uninterrupted token delivery that maintains the illusion of continuous streaming for end users. 

% motivation and insight
We make the analogy between ``LLM text streaming'' and standard video streaming by establishing the request token buffer model, and seeking to serve each request through a ``just-in-time'' manner: Match the average token generation rate for each request with its corresponding token consumption rate and dynamically perform preemptive scheduling between requests upon congestion. 
Specifically, early-arrived requests with high buffer occupancy can be temporarily switched out to serve later-arrived requests first.
Upon request bursts, the reserved GPU computation resources can serve later-arrived requests with lower TTFTs without delaying content consumption for earlier requests. 
The overall request processing parallelism is therefore improved.

% To address these challenges, we propose fundamental innovations in request scheduling and memory management. 
% From computation scheduling, we leverage the inherent mismatch between the system's generation speed and users' consumption rates. 
% Our solution employs client-side token buffering coupled with novel buffer-aware scheduling - the scheduler dynamically monitors and utilizes buffer occupancy levels to make intelligent request prioritization decisions. 
% For memory management, we shift from conventional reactive eviction approaches by introducing a proactive KV cache offloading strategy to CPU memory. 
% Unlike prior systems that perform evictions only when memory pressure occurs, our design maintains ready-to-evict cache copies during idle cycles and employs optimized I/O strategies to minimize overhead during preemption (storing KV cache) and resumption (loading KV cache). This proves particularly effective in handling the irregular preemption/load patterns generated by our scheduler, achieving smooth operation even during request bursts.

% our approach and contributions
We present \method, an optimized LLM serving system that transforms scheduling and memory management for text streaming scenarios upon request bursts. 
Building on SGLang's infrastructure, we introduce a buffer-aware scheduler that dynamically adjusts request priorities based on real-time token buffer states and output rates, enabling intelligent resource allocation without disrupting user experience. 
Instead of maximizing the overall token generation throughput, which may contain substantial ineffective tokens beyond the user comprehension limit, we shift the scheduling objective to maximize the effective throughput that falls within the user consumption rate.
To further handle the memory bottleneck caused by increased request process parallelism, we design a KV cache management module that proactively transfers KV cache entries between GPU memory and CPU memory. To hide memory transfer overhead, it overlaps I/O with computation through background cache preparation and optimized data transfers. 

\method's key innovation lies in transparently coordinated operations between the above two components: The scheduler makes preemption decisions aware of the current I/O load, while the memory manager anticipates scheduling needs to minimize transition overhead. 
This tight integration allows \method to leverage the natural buffer opportunities created by generation-consumption rate mismatches, delivering superior streaming efficiency without additional hardware requirements.

% evaluation 
We evaluate \method across multiple scenarios, including BurstGPT~\cite{wang2024burstgpt} traces, industrial serving traces, and stress tests on RTX 4090, A6000, and H200 GPUs with models from Llama3-8B to Qwen2.5-32B. 
The results demonstrate \method consistently achieves up to $82.5\%$ higher effective throughput (accounting for actual user consumption rates) and reduces time-to-first-token by up to $80.2\%$, while sustaining comparable overall throughput to state-of-the-art (SOTA) baselines. 
These advances establish \method as a robust solution for user-centric LLM streaming, capable of delivering both system efficiency and consistent quality of experience across varying workload patterns and hardware configurations.

Our contributions can be summarized as follows:
\begin{itemize}[topsep=0pt,leftmargin=0.35cm]
    \item \textbf{Streaming QoS metric}: A synthetic metric balancing token usefulness, TTFT, and rebuffering penalties.
    \item \textbf{Buffer-aware scheduling}: A novel scheduling algorithm that dynamically prioritizes requests based on their buffer levels and consumption rates, enabling proactive preemption when generation outpaces consumption (§\ref{sec:4-scheduler}).
    \item \textbf{Hierarchical memory management}: A write-through KV cache system with synchronous chunked writing and load-evict overlap, reducing preemption overhead by 20.3\% (§\ref{sec:5-memory-management}).
    \item \textbf{Consistent responsiveness improvement}: \method achieves up to $82.5\%$ improvements in effective throughput and $80.2\%$ in TTFT.
\end{itemize}

%% file: text/2_background.tex
\section{Background and Motivation}\label{sec:2_background}
% In this section, we first introduce the background knowledge of LLM inference and LLM serving (§2.1), followed by our motivation, the opportunities and challenges of LLM serving in text-streaming scenario (§2.2). 
% We finally discuss the limitations of existing LLM serving systems (§2.3) and KV cache management systems (§2.4).

\subsection{LLM Inference and KV Cache}
% \shengzhong{Keep this subsection short and concise.}

% 1. Two different phase in LLM inference (Prefill and Decode)

% 2. Introduce KV Cache, reuse the key and value to accelerate the inference
Modern LLM 
inference~\cite{sheng2023flexgen,song2024powerinfer,kwon2023efficient,zheng2024sglang,li2023speed,hong2024flashdecoding++,aminabadi2022deepspeed} consists of two distinct phases: (1) the \textit{prefill} phase, where the model processes input prompts of requests to generate initial hidden states; (2) the \textit{decode} phase, where tokens are generated autoregressively. The prefill phase exhibits parallel computation patterns while the decode phase is inherently sequential due to the token-by-token generation nature.
To optimize the decode phase, the KV cache mechanism~\cite{pope2023efficiently} stores attention key-value pairs from previous token generation iterations. By reusing cached KV values instead of recomputing them, the attention computation overhead during generation is significantly reduced, typically achieving a 2-3x speedup in inference throughput. 
% However, this optimization may induce excessive memory overhead as the KV cache grows linearly with the generated sequence length.
However, this optimization comes at the cost of increasing memory bandwidth consumption as the KV cache grows linearly with sequence length.

\subsection{Text-Streaming in LLM Serving}
\begin{figure}[t!]
    \centering
    \includegraphics[width=1\linewidth]{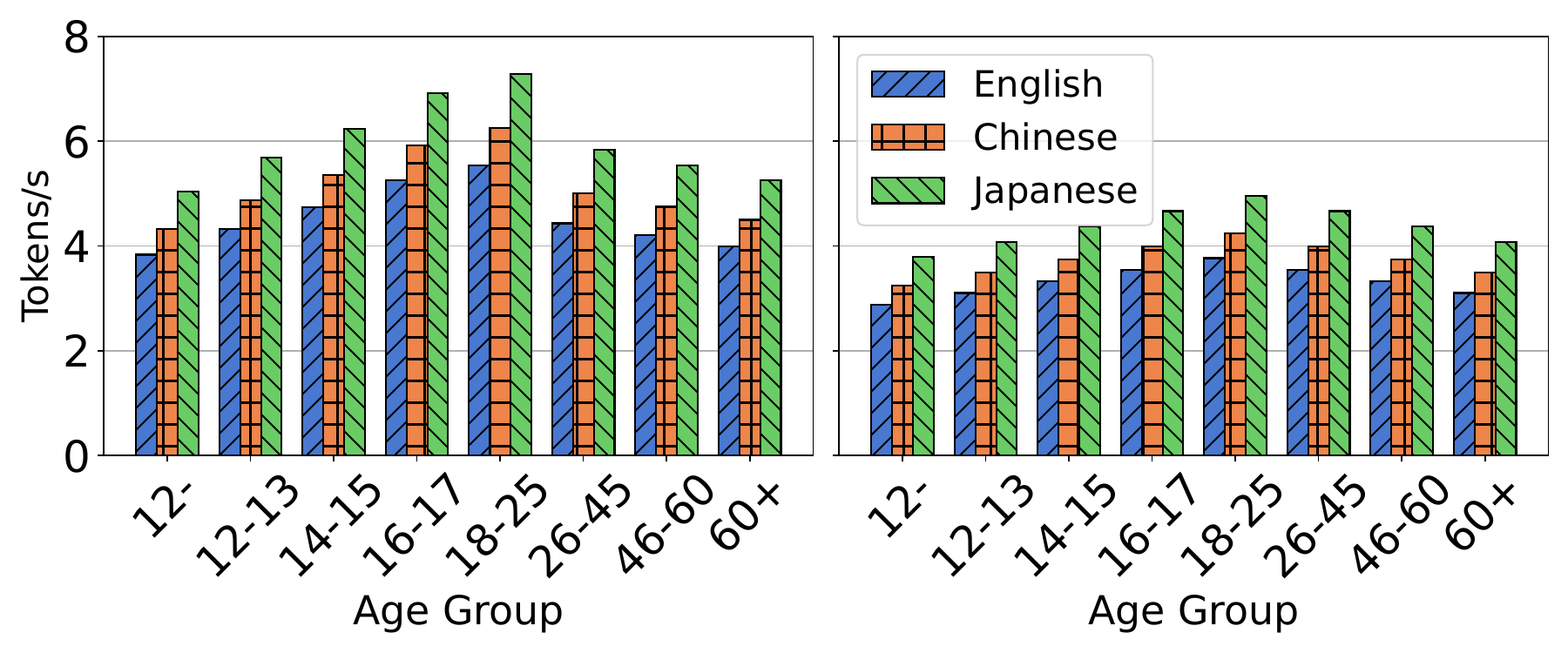}
    \caption{We summarize the token consumption speeds for reading (left) and for listening (right), measured across different age groups and language users. {The data is derived from calculations based on reading speed data from NIH~\cite{liu2017age} and information on token counting from OpenAI's blog~\cite{openai_tokens}}.}
    \label{fig:2-2-different_speed_distribution.pdf}
\end{figure}

% 1. Mismatch between the server generation speed and user reading speed, existing serving systems provide poor user experience when the system is busy (with many reqeusts)
% 2. By introducing the buffer, it is possible for the serving system handle these scenarios (above the GPU memory size) by offloading techniques, to improve the overall performance.
% 思路：这一章的内容是机会和挑战，那么首先说明机会：首先说明LLM文本流式输出的应用场景非常广泛，然后再说LLM文本流式输出的特点，请求之前的异构性很强，这不仅仅是因为请求的输入输出长度不同，请求的来源有可能来自不同年龄段、不同语言，甚至有可能请求的来源不一定是人，可能是AI agent，他们对输出速度的要求也不同，他们接受信息的途径也不同，可能是听也可能是读，也有可能是一个后台任务对速度不敏感。下面是挑战，首先说LLM文本流式输出的两个主要痛点，分别是首字延时对应用户首次收到回复的时间；然后是在生成阶段的卡顿问题，如果输出的速度一旦小于用户阅读的速度，那么就会导致卡顿，而且速度需要稳定，不能给用户带来感知上的速度波动。然后在现有的serving system中对于大量的并发请求的涌入难以做到实时的处理，这也是一个挑战。
\textit{Text-streaming} refers to a progressive token generation and delivery process in LLM decoding, which fits extensive application scenarios, including chatbots, real-time translation, and AI-assisted development tools. 
%However, streaming workloads exhibit inherent request heterogeneity, stemming not only from varying input/output lengths but also from user demographics (\eg age, language proficiency), consumption modalities (\eg reading vs. auditory playback), and request sources (\eg human users vs. AI agents). 
Text streaming exhibits two key characteristics: (1) Content consumers exhibit diverse token consumption rates, and (2) Token consumption rates are typically lower than the LLM's token generation speed.
Analysis across age groups and language backgrounds (Figure~\ref{fig:2-2-different_speed_distribution.pdf}) reveals significant variations in consumption speed, with both reading and listening modes generally consuming tokens much slower than LLMs generate them.
% Analysis of token consumption rates across age groups and language backgrounds (Figure~\ref{fig:2-2-different_speed_distribution.pdf}) reveals that both reading and listening modes consume significantly fewer tokens than the generation rate of standard LLMs.
% 
% This diversity and gap present optimization opportunities in system design. 
These discrepancies lead to differing system requirements across applications.
For example, conversational agents prioritize low initial latency, while real-time captioning requires consistently high generation speed to keep pace with speech. 
Existing rate-agnostic LLM serving systems cannot accommodate such diverse and nuanced demands.

Despite its opportunities, streaming LLM serving must satisfy stringent performance requirements. Empirical studies~\cite{wang2024understanding} reveal that users experience interruptions when generation rates drop below consumption thresholds (\eg <12 tokens/s for reading), while speed variations that exceed 30\% degrade perceived fluency. Furthermore, user engagement suffers when responses are delayed beyond 1.3 seconds.
% [].
% a threshold analogous to Google's recommended web page loading time.
% 
The fundamental challenge stems from simultaneously optimizing two competing objectives: \textit{low initial response time} and \textit{high steady-state generation speed}. This creates inherent resource conflicts, as new requests demand intensive computation during prefill while ongoing streams require stable decode-phase throughput. 
Existing systems struggle to balance both targets under workload spikes, which are further compounded by scheduling inefficiencies under heterogeneous workloads. Overcoming these limitations necessitates fundamental innovations in resource allocation and scheduling policies within LLM serving systems.

% \shengzhong{Streaming is one of our real motivations, but we need to spend more effort in highlighting the limited user comprehension speed, the mismatch with existing schedulers, and why it gives us space for optimization in scheduling. We do not need to introduce the technical details yet, but keep the discussion high-level.}

\subsection{Resource Scheduling for LLM Text Streaming}
% 主要分为两个方面，首先从现有的Scheduler说起，目前的主流的LLM service中的scheduler无论是否有特殊优化，主要都是prefill优先的调度器，他们主要采用FCFS服务策略，这就会带来Head of line延迟，如果请求到来的多，无论请求的要求的输出速度如何，都会在等待，不会因为生成速率超过了阅读速率，就考虑调度其他尚未服务的请求到来，他们就完全没有考虑text streaming场景带来的调度空间。他们的preemption都是被动的preemption，只有当显存空间充满了之后才会进行preemption，而不会根据系统的当前状况，当前的请求要求是否被过度满足，因此他们的preemption也具有滞后性；另一方面是指标方面，现有的吞吐量、TTFT和ITL等孤立的指标无法完整的评估LLM text-streaming的效果
Existing LLM serving systems exhibit significant shortcomings when handling text streaming workloads. 
The prevailing approach adopted by systems relies on first-come-first-served (FCFS) scheduling with priority given to the prefill phase. While this design may optimize for throughput, it remains ill-suited for user-facing interactive applications where latency and smooth token delivery are paramount.

\begin{figure}[t!]
    \centering
    \includegraphics[width=\linewidth]{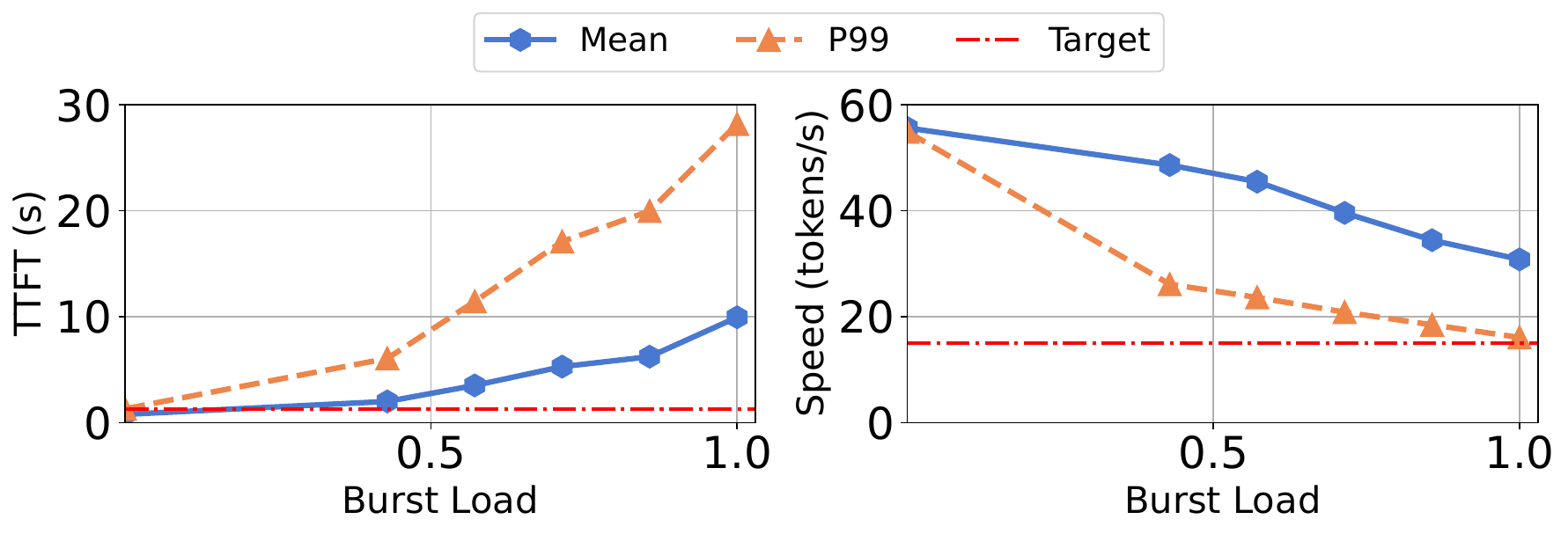}
    \caption{{Micro-benchmark on SGLang's burst request handling conducted on the single NVIDIA H200 GPU.} Left: Time-to-First-Token (TTFT) surges beyond acceptable thresholds (1.3s, red line) under increasing request intensity. Right: Generation speed declines but remains excessively high ($2\times$ average reading speed for reference, red line).}
    \label{fig:2-3-current-serving-system}
\end{figure}

Our micro-benchmark reveals a critical mismatch between text streaming demands and existing LLM scheduling. As demonstrated in Figure~\ref{fig:2-3-current-serving-system}, request surges create substantial queuing delays that severely degrade user-perceived latency. 
Real-world trace analysis of SGLang shows peak-load time-to-first-token (TTFT) frequently exceeding 20 seconds, far beyond acceptable user tolerance limits. Paradoxically, while some requests experience excessive queueing delays, active-served requests achieve unnecessarily high generation speeds (averaging 30 tokens/second in our tests). 
This resource misallocation provides no practical benefit in consideration of practical user token comprehension speed.
% , as typical user token consumption rates remain significantly lower than these maximum throughput levels.

The fundamental mismatch stems from limitations in current scheduling mechanisms. First, preemption is used solely as a passive memory management strategy, rather than as a proactive tool for optimizing resource allocation.
% First, existing systems employ preemption only as a passive memory management strategy rather than proactively optimizing resource allocation. 
Second, although excess token generation can create buffers that would allow non-disruptive request eviction, existing methods fail to take advantage of this opportunity.
% Second, while accumulated buffers from excess token generation could enable non-disruptive request eviction, current methods fail to leverage this opportunity. 
As a result, these systems struggle to adapt dynamically to real-time demand fluctuations and fall short in leveraging the relationship between generation and consumption rates (\ie they lack true \textit{buffer-awareness}). 
% The existing approach demonstrates an inability to dynamically adjust to real-time demand fluctuations and insufficient exploitation of the generation-consumption rate relations (\ie \textit{buffer-awareness}). 
% Consequently, systems miss opportunities to improve efficiency without degradation of service.

The current evaluation methodology for streaming text generation presents another limitation:
it relies heavily on conventional metrics such as throughput, Time-To-First-Token (TTFT), and inter-token latency, each of which captures only a narrow aspect of performance. 
% which relies on conventional metrics like throughput, Time-To-First-Token (TTFT), and inter-token latency, each offering only a partial performance assessment. 
These metrics focus on isolated system behaviors and fail to reflect the overall user experience.
For example, prefill-centric scheduling strategies may optimize throughput but often cause unacceptable increases in TTFT during request surges. This highlights why no single metric can adequately evaluate streaming services in isolation.
% For instance, current prefill-centric scheduling prioritizes throughput and may lead to unacceptable TTFT increases during request surges.
% It explains why text streaming services cannot be properly evaluated using any single metric in isolation.
Recent work like Andes' QoE metric~\cite{liu2024andes} improves user experience evaluation but overlooks system efficiency. 
An effective framework must balance both user-perceived experience and computational resource utilization for optimal streaming service.

\subsection{Hierarchical Memory Management for LLM}
% 现有的内存管理系统并不能很好的为preemption机制的LLM serving system提供显存管理的服务。如果在显存占用过大的情况下才进行显存的offload或者是压缩等操作，会导致延迟，使得系统的瓶颈变为IO瓶颈。另外现在的显存管理系统主要是一个单一的模块，没有和调度系统和推理内核形成联动，现有的工作如Andes仅仅优化了调度逻辑，而并没有结合显存管理模块一同进行优化，这导致了他们次优的结果
Current memory management policies fall short in supporting effective preemption within LLM serving architectures. The core limitation lies in their reactive design: memory offloading is only triggered when GPU utilization reaches a critical threshold. This delayed response incurs significant I/O overhead, shifting the system from compute-bound to I/O-bound.
% Current memory management policies fail to adequately support preemption in LLM serving architectures. The fundamental limitation lies in their reactive nature - memory offloading or compression is only triggered when GPU memory utilization reaches a critical threshold, inevitably introducing substantial I/O overhead and transforming a compute-bound system into an I/O-bound one.
% 
% The architectural shortcomings become particularly apparent when examining the disjointed design of contemporary systems. 
Moreover, memory management operates in isolation from both the scheduling system and the inference engine, leading to poor coordination and suboptimal decisions across the stack. For example, while Andes has shown performance gains through improved scheduling logic, its full potential remains unrealized without co-designed memory management. 
% The disconnect between these subsystems results in local optimizations that fall short of overall system efficiency.
% Memory management operates as an isolated module in parallel to the scheduling system and inference kernel. This lack of coordination results in suboptimal decision-making at each stack layer. For instance, Andes has demonstrated improvements through enhanced scheduling logic, but their potential remains under-exploited without co-designed memory management components. The separation of concerns between these critical subsystems leads to local optimizations that fail to achieve global efficiency.
This lack of integration leads to local optimizations that undermine overall system efficiency. It manifests in three key shortcomings: (1) the memory manager lacks visibility into the scheduler’s preemption decisions, forcing it to evict based solely on memory pressure rather than scheduling intent; (2) the inference engine cannot anticipate memory operations, leading to unpredictable runtime latency spikes; and (3) the absence of a unified management strategy results in redundant or conflicting actions, wasting precious memory bandwidth and compute cycles.
% Three concrete limitations are present: 
% First, the memory manager lacks visibility into the scheduler's preemption decisions, forcing it to make eviction decisions based solely on memory pressure rather than actual scheduling needs. 
% Second, the inference engine cannot anticipate memory management operations, resulting in unpredictable runtime latency spikes. 
% Third, the absence of a unified management strategy leads to redundant or conflicting operations that waste precious memory bandwidth and compute cycles.

%% file: text/3_overview.tex
\section{Overview and Formulation} \label{sec:3_framework}
% In this section, we first give an overview of the proposed framework (§3.1), then we will introduce the overall problem formulation of our system (§3.2, §3.3).

\subsection{\method Overview}

\begin{figure*}[t!]
    \centering
    \includegraphics[width=0.9\linewidth]{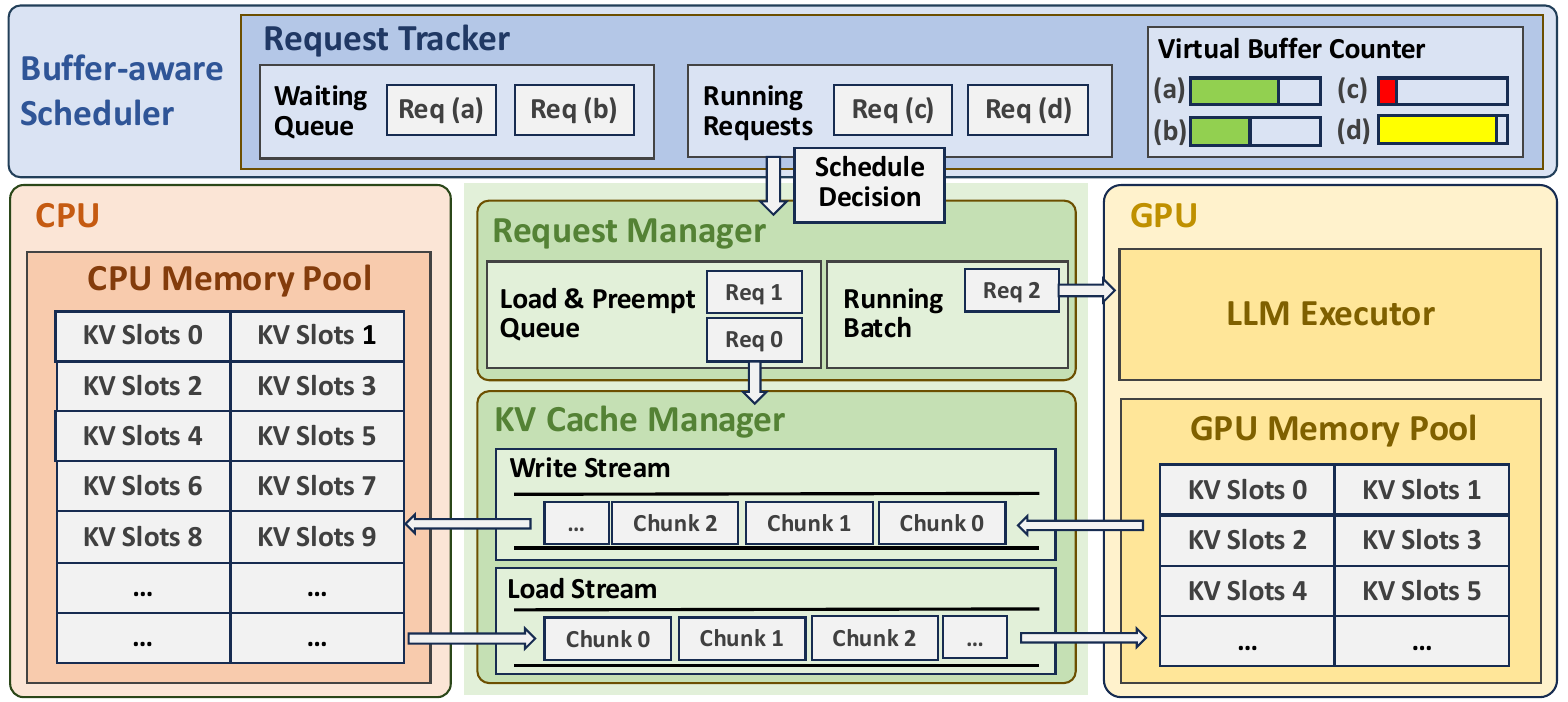}
    \caption{Overview of \method: Detailed breakdown of all modules and their components.}
    \label{fig:3-1-overview}
\end{figure*}

\mypara{System Components}
% \method presents a co-designed architecture that integrates the LLM inference server with its KV cache management system, as shown in Figure~\ref{fig:3-1-overview}. The system consists of five key components working in coordination. This integrated architecture enables \method to achieve efficient resource utilization while meeting the latency and throughput requirements of text streaming applications.
As shown in Figure~\ref{fig:3-1-overview}, \method introduces a co-designed architecture that tightly integrates a preemptive LLM scheduler with proactive KV cache management. The system's five cooperating components work together to maximize resource utilization while balancing the latency and throughput demands for requests.
\begin{itemize}[topsep=0pt,leftmargin=0.35cm]
\item \textbf{Request Tracker} monitors each request's status, including buffer token counts, latency targets, user consumption rates, prompt/response data, token generation timestamps, and resource usage.
\item \textbf{Buffer-aware Request Scheduler} accepts real-time metrics and implements the scheduling algorithm(§3.2) to make runtime decisions about request admission, preemption, and resumption. 
\item \textbf{Request Offload Manager} executes the scheduler's decisions by managing request-level memory operations, evicting requests via a write queue, and restoring them via a loading queue, thus bridging high-level scheduling with low-level execution.
\item \textbf{LLM Executor} is built primarily upon SGLang with minimal modifications and performs the actual LLM inference and token generation for active requests. 
\item \textbf{Hierarchical KV Cache Manager} efficiently manages token-level memory operations across separate CPU and GPU pools. Its specialized writing and loading queues handle token chunks, minimizing I/O overhead while maintaining performance.
\end{itemize}

\begin{figure}
    \centering
    \includegraphics[width=1\linewidth]{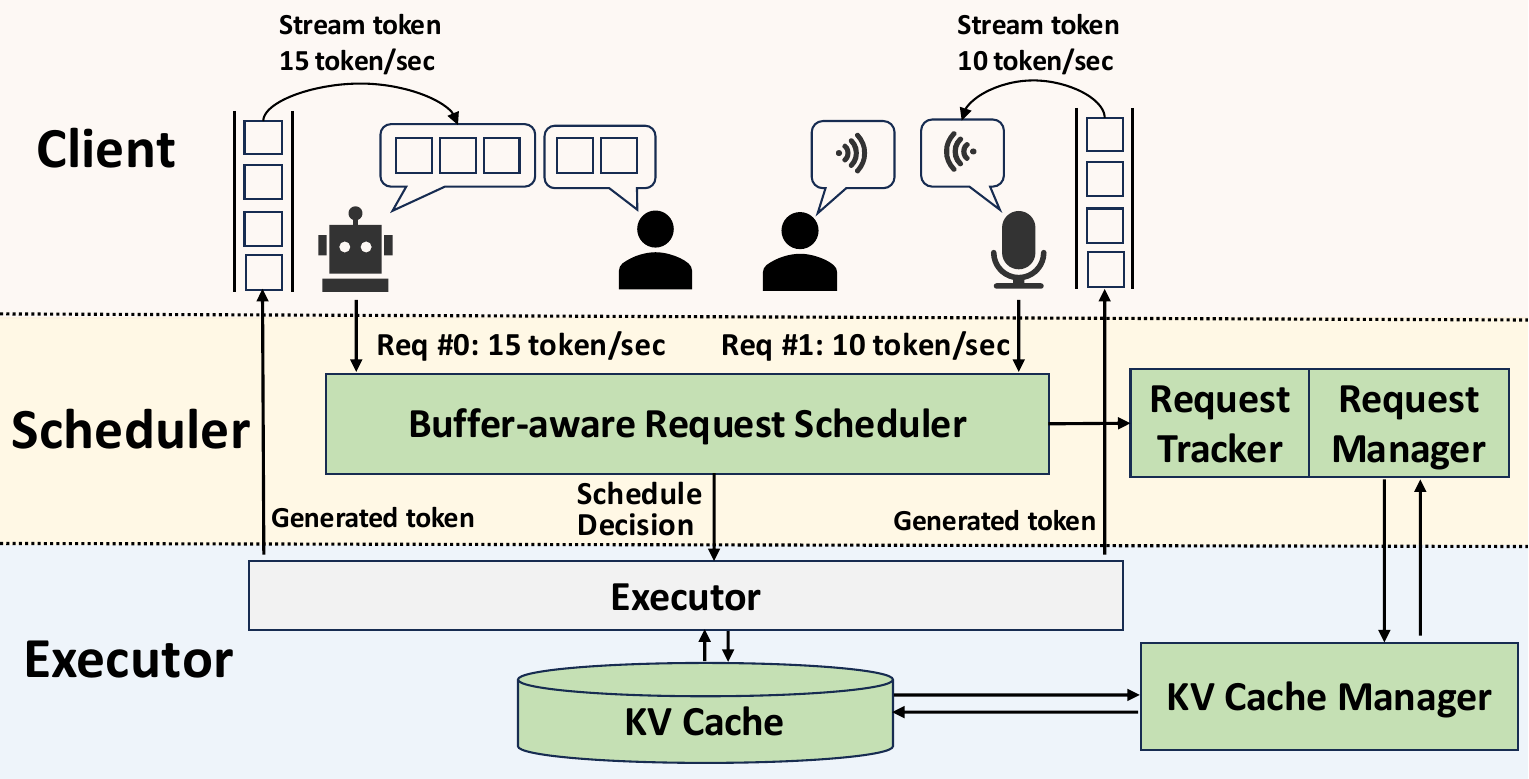}
    \caption{High-level workflow of \method. Modules newly added by \method are colored green.}
    \label{fig:3-3-workflow}
\end{figure}

\mypara{Workflow}
The coordinated workflow is illustrated in Figure~\ref{fig:3-3-workflow}.
Each request undergoes a carefully managed lifecycle in our system. 
When a client (\eg chatbot or voice assistant) submits a request, it arrives at the server carrying the streaming speed requirement and maintains a client-side token buffer. 
The Request Tracker registers the request and begins monitoring its state transitions between waiting (while enqueued or preempted) and active execution phases. 
The Buffer-aware Scheduler dynamically controls the progress by preempting running requests after preserving their state through coordinated KV cache management or activating waiting requests through admission-resumption iterations after loading their state. 
During execution, the LLM Executor generates and streams tokens directly to the client's buffer, which then paces delivery according to the given consumption rate, while the system continuously optimizes compute and memory resource allocation among requests to maintain quality-of-service (QoS) guarantees.

%The process begins when a user submits a request through an application-integrated client (\eg chatbot or voice assistant), which specifies the desired streaming speed and maintains a token buffer for excess generation. Upon arrival at the server, the Request Tracker initializes and continuously monitors the request's lifecycle state, alternating between waiting (enqueued/preempted) and running phases.

% The Buffer-aware Request Scheduler dynamically manages execution through two key operations: (1) preempting running requests by transitioning them to waiting state after state preservation (handled by the Request Manager and KV Cache Manager), and (2) activating waiting requests either through new admission or resumption (where the Executor restores intermediate states from Hierarchical KV Cache when available). During execution, the Executor streams generated tokens directly to the client's token buffer, which then regulates delivery at the specified consumption rate.

\subsection{Quality of Text-Streaming Service Metric}
Standard throughput metrics (\eg tokens/second) inadequately measure user experience factors like responsiveness and streaming smoothness.
We therefore define a comprehensive \textbf{Quality of Service (QoS)} metric that integrates the token utility values, first-token delays, and penalties for playback stalls. 
Unlike throughput, our QoS metric better reflects actual user experience by evaluating initial delay, delivery consistency, and buffer efficiency, while simultaneously considering system responsiveness and resource utilization.
QoS optimization thus achieves a better balance between minimizing initial latency to the first response and maximizing processing efficiency for enhanced text streaming.

\begin{figure}[t!]
    \centering
    \includegraphics[width=1\linewidth]{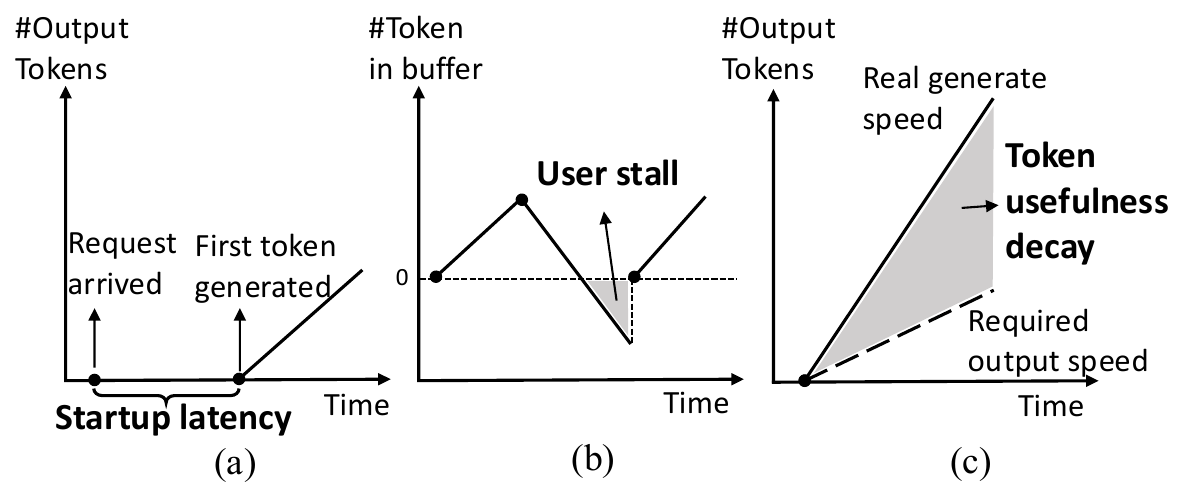}
    \caption{Three QoS factors: (a) Startup latency, (b) User stall events, (c) Token usefulness.}
    \label{fig:3-2-qos}
\end{figure}

Assume the system handles a batch of $N$ requests, where each request $i$ is characterized by:
\begin{itemize}[topsep=0pt,leftmargin=0.35cm]
    \item Time-to-first-token (TTFT), denoted as $t_i^{\text{ttft}}$,
    \item A sequence of inter-token latencies $\{\delta_{i,1}, \delta_{i,2}, \dots, \delta_{i,L_i}\}$, where $L_i$ is the number of generated tokens,
    \item A fixed user reading speed $r_i$ in tokens/second.
\end{itemize}

The user starts reading at time $t_i^{\text{ttft}}$, consuming one token every $1/r_i$ seconds, and  token $j$ of request $i$ is generated at time:
$t_{i,j}^{\text{gen}} = t_i^{\text{ttft}} + \sum_{k=1}^{j-1} \delta_{i,k}.$
Let $B_{i,j}$ denote the size of the output buffer for request $i$ at the moment token $j$ is generated. We assign each token a weight $w_{i,j} \in [0, 1]$ representing its utility, defining this as either the \textit{token utility} or the \textit{token weight}:
\begin{equation}
    w_{i,j} = 
    \begin{cases}
        1, & \text{if } B_{i,j} \leq \tau \\
        \max(1 - \alpha \cdot (B_{i,j} - \tau), 0), & \text{if } B_{i,j} > \tau
    \end{cases}
\end{equation}
where $\tau$ is the buffer threshold beyond which usability starts to decay, which is related to the total output length of the request, and $\alpha > 0$ is a tunable decay factor.
Let $\text{Rebuffer}_i$ denote the total time user $i$ experiences an empty buffer when reading a token, and $t_i^{\text{ttft}}$ denote TTFT of request $i$. We apply penalties to both quantities, with weight coefficients $\lambda$ and $\mu$ reflecting the importance of low latency and uninterrupted experience.
Finally, we define the \textbf{Quality of Service (QoS)} of the text-streaming LLM serving system as:
\begin{equation}
    \text{QoS} = \frac{1}{T} \sum_{i=1}^{N} \left( \sum_{j=1}^{L_i} w_{i,j} - \lambda \cdot t_i^{\text{ttft}} - \mu \cdot \text{Rebuffer}_i \right),
\end{equation}
where $T$ is the overall request process time. 

This metric provides a more accurate evaluation of inference scheduling policies and system responsiveness than raw throughput alone. 
As shown in Figure~\ref{fig:3-2-qos}, the QoS metric accounts for:
\begin{itemize}[topsep=0pt,leftmargin=0.35cm]
    \item \textbf{Token usefulness}: Only tokens that are timely and within buffer limits contribute fully.
    \item \textbf{Startup latency}: Delays in first-token generation are penalized.
    \item \textbf{User stall}: Gaps in token availability during playback degrade experience and reduce QoS.
\end{itemize}

\subsection{Scheduling Problem Formulation}
We first discuss the overall scheduling objective and associated constraints of our methods, and then formally formulate the scheduling problem.

\mypara{Objective} 
While QoS effectively measures text-streaming quality, it presents practical challenges in online system optimizations. The metric requires complete request traces, including inner-token latencies and rebuffering durations, which are only available after request completion. 
However, real-world systems must handle dynamic request arrivals with unpredictable characteristics (\ie arrival times, lengths, and output speeds), making direct QoS optimization during live scheduling infeasible.

% To address this, we propose a practical proxy objective that aligns with QoS while being tractable in online scheduling. At each scheduling step, we aim to maximize the expected number of useful tokens that can be generated by the selected set of requests. Here, a token is considered useful if it is likely to be read soon by the user and will not cause the output buffer to exceed a usable threshold. Additionally, we introduce a penalty term to discourage schedules that result in underfilled buffers, as it may lead to reduced user-perceived quality, indicate risk of playback stalls.

To enable practical QoS optimization during online scheduling, we propose a tractable proxy objective that maximizes the expected number of generated effective tokens, those likely to be consumed promptly without exceeding buffer capacity, while penalizing schedules that risk buffer underflows and potential playback stalls.
To further refine scheduling decisions at the granularity of individual requests, we define a per-request utility function that balances the value of token generation and the buffer state, then the total utility of selected requests $\sum\mathcal{U}_i$ is the tractable objective.
Given a request $i$, its utility function is defined as:
\begin{equation}
\mathcal{U}_i = v_i \cdot t - \gamma \cdot \phi(b_i^{\text{rem}}),
\end{equation}
where $t$ is the allocated execution time for this request in the current scheduling step,
$b_i^{\text{rem}}$ is the number of unread tokens in its output buffer,
$v_i$ is the estimated token value, related to the unread tokens number $b_i^{\text{rem}}$ in buffer,
$\phi(\cdot)$ is a penalty function that increases when the buffer is too low (risking stall),
$\gamma$ is a tunable regularization coefficient controlling the penalty strength.

\mypara{Resource Constraint} % Memory, Prefill/Decode time, Buffer size, KV transmission time
LLM streaming systems face two fundamental constraints: \textit{GPU memory limits} and \textit{batch-size-dependent computational trade-offs}.
The first constraint is familiar to LLM scheduling: KV cache memory bounds the number of concurrent requests under service. Each active request occupies fixed memory for context tokens, capping system parallelism.
The second constraint is unique to streaming scenarios: batch size $B$ critically impacts both performance and scheduling stability. The core challenge is to optimize $B$ dynamically, balancing memory use, I/O overhead, and decode throughput while preventing buffer underflows. Unlike traditional systems with a restricted focus on throughput, streaming introduces delicate dynamics:
\begin{itemize}[topsep=0pt,leftmargin=0.35cm]
    \item \textbf{Batch Size vs. I/O Overhead:} Significant batch size variations force either (1) new requests allocating fresh KV cache, or (2) reloading evicted requests from the CPU memory. Both incur I/O latency that risks depleting buffers during stalled computation.
    \item \textbf{Batch Size vs. Decode Speed:} Large batches saturate memory bandwidth, slowing token generation and buffer accumulation, while small batches improve responsiveness but waste GPU computation. This creates tension between preemption flexibility and hardware utilization.
\end{itemize}

\mypara{Problem Formulation} 
% Given the definition of Effective Throughput (ET), we aim to formulate a scheduling optimization problem that selects the optimal subset of requests to serve and determines their processing order to maximize ET under system constraints such as KV cache size, concurrent request limits, and prefill capacity.
We formulate LLM request scheduling as an online combinatorial optimization problem that selects request subsets at each interval $t$ to maximize utility while respecting hardware constraints, as shown below:
\begin{align*}
\max_{x} \quad & \sum_{i \in \mathcal{A}} x_i \cdot \left(v_i(t - t_i^{\text{overhead}}) - \gamma \phi(b_i^{\text{pred}})\right), \\
\text{s.t.} \quad & x_i \in \{0,1\} \quad \forall i \in \mathcal{A}, \\
& \sum x_i \leq B, \\
& \sum x_i l_i \leq, M
\end{align*}
where $\mathcal{A}$ denotes the set of all active requests at the scheduling  moment. Each request $i \in \mathcal{A}_t$ is associated with a binary variable $x_i \in \{0, 1\}$, indicating whether it is scheduled in this step ($x_i = 1$ if $i \in \mathcal{S}_t$).
$B$ is the maximum number of concurrent running requests allowed by the system. Each request $i$ has a context length $l_i$, which determines its key-value cache memory footprint. $M$ is the total available GPU memory. 

% There are two key modifications to the optimization problem: replacing the original time allocation $t$ with the effective execution time $t - t_i^{\text{overhead}}$ and substituting the current buffer state $b_i^{\text{rem}}$ with the predicted buffer state $b_i^{\text{pred}}$.
% The context-switch overhead $t_i^{\text{overhead}}$ captures the latency from memory operations including offloading, reloading, and recomputation, which varies based on scheduling decisions and available I/O bandwidth. The predicted buffer state $b_i^{\text{pred}}$ estimates future unread tokens after considering these system overheads.
The optimization problem also incorporates two key refinements: (1) using the effective execution time $t - t_i^{\text{overhead}}$ instead of $t$ accounting for context-switch latency of memory operations and
(2) employing predicted buffer states $b_i^{\text{pred}}$ that anticipate token accumulation under system overhead. {Here, $t_i^{\text{overhead}}$ (context switching time) is estimated as $\min(t_\text{IO}, t_\text{recompute})$. In particular, $t_\text{IO}$ is obtained from the memory manager’s profiled memory read/write throughput based on execution history, and $t_\text{recompute}$ approximates prefill time using sliding-window-averaged per-token latencies (details given in Section~\ref{sec:4-scheduler}). In short, since these estimates are inherently prompt-dependent, we incorporate both historical execution records and characteristics of the user-provided prompt to derive such metrics.}
These modifications explicitly model the performance impact of memory management and I/O constraints during request processing scheduling.

% This formulation captures the balance between utility-driven request selection and hardware constraints. Though the optimization is combinatorial and must be solved online, it provides a principled target for heuristic scheduling policies.

% \begin{itemize}
%     \item $\text{ttft}_i$: estimated time-to-first-token,
%     \item $\text{ITL}_i = \{\delta_{i,1}, \dots, \delta_{i,L_i}\}$: inter-token latency sequence,
%     \item $r_i$: user token reading speed,
%     \item $k_i$: KV memory cost (in tokens),
%     \item $p_i$: whether the request is a prefill ($p_i = 1$) or decode-only ($p_i = 0$).
% \end{itemize}

% Our objective is to maximize the total Effective Throughput across selected requests:

% \begin{equation}
% \max_{x_1, \dots, x_M} \quad \frac{1}{T} \sum_{i=1}^{M} x_i \left( \sum_{j=1}^{L_i} w_{i,j} - \lambda \cdot \text{ttft}_i - \mu \cdot \text{Rebuffer}_i \right)
% \end{equation}

% % Where $w_{i,j}$ and $\text{Rebuffer}_i$ depend implicitly on the token generation and scheduling order, and may be estimated via simulation or approximated by heuristics.

%% file: text/4_scheduler.tex
\section{Buffer-Aware Request Scheduling}
\label{sec:4-scheduler}
% Briefly introduce the organized paragraph here:

% 1. Introduce our solution (buffer aware scheduler)

% 2. (Optional? in this part or in experiment part) Discuss the overhead of the scheduler

% Explain the difficulty to solve the problem in linear time, we use an approximate solution by greedy search algorithm

The scheduling problem formulated in the previous section is a combinatorial optimization problem with discrete decisions and nonlinear execution cost, making exact solutions computationally intractable in real-time serving scenarios. 
To address this, we design a heuristic buffer-aware scheduling algorithm that approximates the original objective by efficiently prioritizing and selecting requests based on their estimated utility and system constraints. 

\subsection{A Motivating Example}
\begin{figure}[t!]
    \centering
    \includegraphics[width=0.98\linewidth]{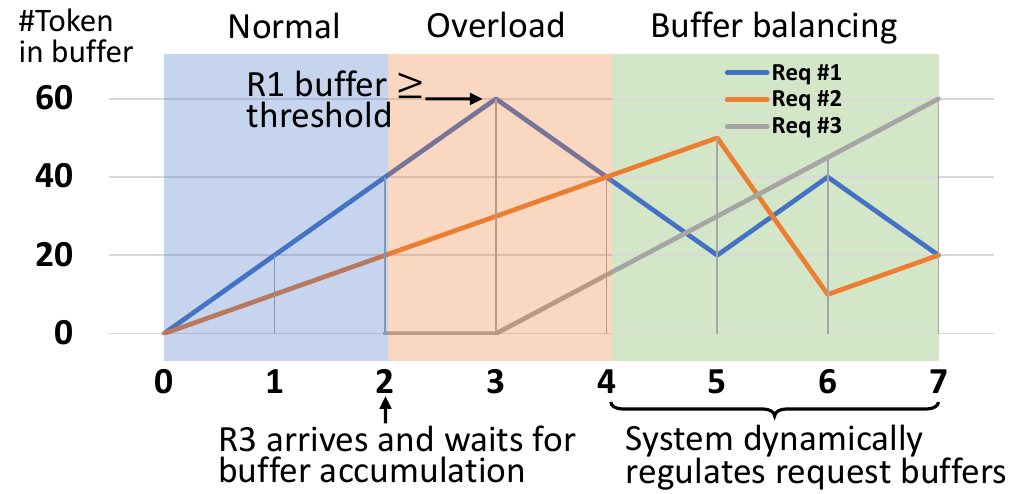}
    \caption{Toy Example of Buffer-Aware Request Scheduling.} %\shengzhong{Shorter.}}
    \label{fig:4-1-toy-example}
\end{figure}

As shown in Figure~\ref{fig:4-1-toy-example}, we illustrate the scheduling process with an operational example to demonstrate the scheduler's behavior. 
Consider a system with a generation capacity of 40 tokens/sec supporting two concurrent requests. Initially, requests R1 (20 tokens/sec) and R2 (30 tokens/sec) arrive at $t=0$, both accumulating surplus tokens at the full system rate. When R3 (25 tokens/sec) arrives at $t=2$, neither active request has sufficient buffer for preemption, demonstrating buffer-dependent admission control. By $t=3$, R1’s lower demand results in sufficient buffer growth, enabling its safe preemption: its context is offloaded to the CPU while tokens continue being served from its buffer, allowing R3 to execute. The system maintains seamless service for R1 via its buffer reserves. At $t=5$, as R1’s buffer nears depletion, the scheduler preempts R2 (now with the largest buffer) to reactivate R1, preventing stalls without sacrificing throughput. Subsequent scheduling decisions similarly balance token reserves across requests, ensuring efficient resource utilization.

Our methodology enables significant system capacity expansion by dynamically prioritizing requests where the utility function of request is maximized, mostly those with smaller buffer sizes. This reveals a key operational insight: \textbf{Maintaining buffers within an optimal range is crucial to improving QoS}. 
However, practical implementation must account for recomputation and I/O overheads that preclude naive empty-buffer scheduling. Effective deployment requires predictive pre-loading and continuous system monitoring to navigate these constraints while pushing capacity limits. 
% These practical considerations, which we explore in subsequent sections, define the boundary conditions for real-world system optimization.

\subsection{Two-Step Scheduler Design}
\begin{figure}[t!]
    \centering
    \includegraphics[width=1\linewidth]{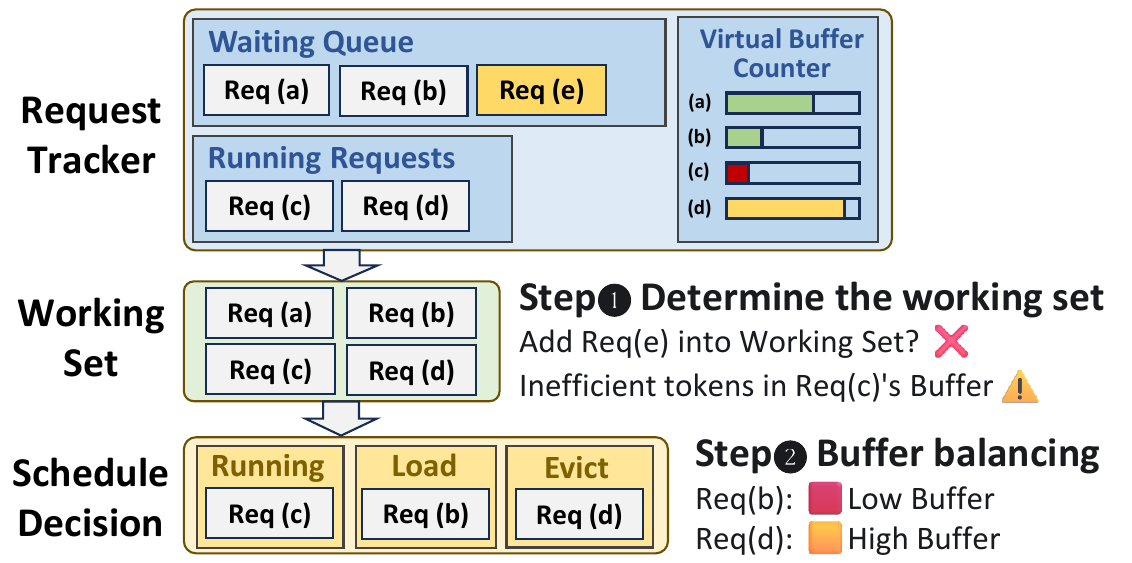}
    \caption{Two-step scheduling in \method: An example workflow showing one complete scheduling cycle of our buffer-aware approach.}
    \label{fig:4-2-two-step-scheduler}
\end{figure}

As illustrated in Figure~\ref{fig:4-2-two-step-scheduler}, the proposed request scheduling algorithm operates in two phases:
\begin{enumerate} [wide, labelwidth=!, labelindent=0pt, label=\protect\blackcircled{\arabic*}]
    \item \textbf{Working Set Determination}: Selecting the set of requests actively processed by the system.
    \item \textbf{Buffer Balancing}: Ensuring equitable buffer allocation across the working set to prevent overflow/underflow.
\end{enumerate}
% This dual-phase approach optimizes resource utilization while maintaining stability during high-load scenarios.

\subsubsection{Determine the working set.} 
The working set defines the upper limit for overcommitment scheduling, based on hardware capabilities and request characteristics. During runtime, we dynamically adjust the working set size by monitoring system conditions, including I/O queue length, the token buffer size and the output rate of each request. A new request is admitted into the working set if the buffers of existing requests meet specific criteria (\eg remaining tokens can be processed within swap/reschedule latency).

The working set size $W$ is determined first by hardware constraints. The static upper bound $W_{\text{static}}$ is computed as:  
\begin{equation}
    W_{\text{static}} = \left\lfloor M / \beta \right\rfloor,
\end{equation}
where $M$ is the total memory capacity, and $\beta$ is the estimated per-request memory footprint. 
% 
% During runtime, the size of the working set \( W \) is dynamically adjusted based on the number of requests currently running \( N_{\text{running}} \). When \( N_{\text{running}} < W_{\text{max}} \), the working set is reduced proportionally to maintain efficiency:
% \begin{equation}
% W_{\text{scheduled}} = W_{\text{static}} - \lambda \cdot (W_{\text{static}} - N_{\text{running}}), 
% \end{equation}
% where \( W_{\text{current}} \) is the current working set size, \( W_{\text{static}} \) is the nominal working set size, \( N_{\text{running}} \) is the number of currently running requests and \( \lambda \) is a scaling factor that controls the rate of adjustment. Otherwise, the working set size \(W\) will not change to keep high efficiency.
During runtime, the working set size \( W \) is dynamically adjusted based on the number of currently running requests \( N_{\text{running}} \). If \( N_{\text{running}} < W_{\text{max}} \), the working set is scaled down proportionally:  
\begin{equation}
    W_{\text{scheduled}} = W_{\text{static}} - \lambda \cdot (W_{\text{static}} - N_{\text{running}}), 
\end{equation}
where \( W_{\text{static}} \) is the nominal working set size, \( \lambda \) controls the adjustment rate, and \( W \) remains unchanged when \( N_{\text{running}} \geq W_{\text{max}} \) to sustain high throughput.  

% Our scheduler operates on a \textit{time-slice-based mechanism}, triggered at regular intervals \(t_{\text{schedule\_interval}}\) to improve efficiency. The scheduler utilizes a demand-driven activation mechanism to minimize scheduling overhead. Specifically, scheduling operations are triggered exclusively upon the detection of either of the following system stress indicators: (1) Pending requests in the queue: \(Q_{\text{waiting}} > 0\); (2) Critical buffer conditions in any active request: \(\exists\; i \;\text{s.t.}\; T_i < T_{\text{critical}}\).

% Under normal operating conditions, the system conservatively applies the default prefill-first policy without explicit scheduling interventions. This design ensures that the scheduling overhead remains strictly proportional to system demand, optimizing computational efficiency during both stressed and steady-state operation.
% The scheduler periodically evaluates and adjusts the working set size W through a time-slice-based mechanism operating at fixed intervals $t_{\text{schedule\_interval}}$. This demand-aware approach minimizes overhead by triggering scheduling operations only when system stress occurs , either when pending requests accumulate $Q_{\text{waiting}} > 0$ or when any active request reaches a buffer-critical state $T_i < T_{\text{critical}}$. During normal operation, the system defaults to an efficient prefill-first policy, ensuring scheduling overhead remains proportional to actual demand while maintaining optimal performance across all operational states.
The scheduler employs a time-sliced mechanism operating at fixed intervals $t_{\text{schedule\_interval}}$ to dynamically manage the size of the working set $W$. This demand-sensitive approach optimizes system efficiency by activating scheduling operations exclusively during stressing periods, characterized by either accumulated pending requests $Q_{\text{waiting}} > 0$ or buffer-critical conditions in active requests $T_i < T_{\text{critical}}$. 
Otherwise, the system maintains optimal performance through a prefill-first policy that scales scheduling overhead proportionally with actual demand.

% A request \( R_i \) is admitted to the working set only when \( W_{\text{current}} < W_{\text{scheduled}} \) and its buffer token count satisfies \( b_i^{\text{rem}} \geq \mu \cdot r_i \cdot (\tau_{\text{evict}} + \tau_{\text{load}} + \tau_{\text{schedule}}) \), where \(\mu \geq 1\) is a safety factor, $r_i$ is the required output speed of request $i$. This conditional admission control ensures system stability by preventing resource over-commitment and maintaining the overall quality of service. 
During each scheduling iteration, the scheduler evaluates whether to admit pending requests into the working set. A request $R_i$ gains admission when the working set has available capacity $W_{\text{current}} < W_{\text{scheduled}}$ and when its remaining buffer tokens meet the requirement $b_i^{\text{rem}} \geq \mu \cdot r_i \cdot (\tau_{\text{evict}} + \tau_{\text{load}} + \tau_{\text{schedule}}) $. Here, $\mu \geq 1$ serves as a safety factor (buffer conservativeness) to account for operational variability, while $r_i$ denotes the request's required output rate. This admission policy prevents resource overcommitment while sustaining consistent service quality.

\subsubsection{Buffer balancing inside the working set.} 
To support more concurrent requests with the given GPU capacity, \method employs an overcommitment mechanism based on the working set model. When the working set exceeds available GPU memory, requests are transparently offloaded to CPU memory. This necessitates an efficient preemptive scheduling strategy to manage the GPU-CPU memory hierarchy.  
Our scheduler dynamically assigns a priority to each request based on the utility function: $\mathcal{U}_i = v_i \cdot t' - \gamma \cdot \phi(b_i^{\text{rem}})$, 
which incorporates the following key factors:  
\begin{enumerate} [wide, labelwidth=!, labelindent=0pt, label=\protect\blackcircled{\arabic*}]
    \item \textbf{Buffer Size ($b_i$)}: The token count currently in the request's buffer. A larger buffer suggests the request can tolerate some delay, reducing its immediate processing priority. We model this effect using an exponential decay function: $\phi(b_i^{\text{rem}}) = e^{-b_i^{\text{rem}}}$, 
   ensuring that requests with nearly empty buffers receive higher priority to prevent starvation.  
    \item \textbf{Weighted Token Generation Quantity ($v_i \cdot t'$)}: The expected number of tokens generated during the schedule interval, weighted by their likelihood. $t'$ approximates the queuing delay of the request, accounting for its position in the batch processing pipeline. We estimate $t'$ using a moving average instead of computing the exact queuing delay from dynamic scheduling.
\end{enumerate}

To select the optimal subset of requests for execution, we employ a greedy algorithm with a local search strategy. Requests are first sorted by their buffer size ($\phi(b_i^{\text{rem}})$), weighted token generation quantity ($v_i \cdot t'$), and their required output rate $r_i$. 
The greedy algorithm then iteratively selects the highest-priority requests that can fit in the available GPU memory, maximizing immediate utility. While this method efficiently provides an approximate solution, it may overlook marginally better request orderings.

Furthermore, we perform a local search by evaluating adjacent request swaps in the priority queue. For each pair of adjacent requests, we compute the potential utility gain from swapping their order. A swap is applied if it improves overall throughput or fairness without violating memory constraints, ensuring the scheduler explores small perturbations to the greedy solution, further optimizing system performance while keeping low computational overhead.  

% This combined approach ensures that the system can handle more requests than its nominal capacity, while minimizing overhead and maximizing throughput, all while maintaining system stability and smooth token delivery.

\subsubsection{Balance recompute and load from CPU memory.}
The scheduler dynamically determines whether to reload offloaded requests from CPU memory or recompute them, balancing faster loading speed against potential head-of-line blocking from loading queueing delays. The I/O overhead is:
$$
t_{\text{IO}} = t_{\text{evict\_queueing}} + t_{\text{evict}} + t_{\text{load\_queueing}} + t_{\text{load}},
$$  
where each term corresponds to the queuing/execution phases during data movement between the GPU and CPU memories. At runtime, \method monitors instantaneous I/O queue lengths and transfer rates via the cache manager.

For prefill and recompute time estimation, \method evaluates recomputation costs using sliding-window-averaged prefill latencies per token. A request is recomputed if $t_{\text{IO}}$ exceeds its estimated recomputation time, thereby adapting to load contention. 
Batching recomputation with new prefill requests risks prolonged memory occupancy, as all allocated blocks must be released before batch completion. This serialization delay directly impacts TTFT for subsequent requests. To mitigate memory contention, the system avoids prolonged block retention by batching carefully and dynamically partitions prefill batches based on remaining capacity and priority. 
It helps latency-sensitive requests bypass batch processing when needed, balancing both throughput and TTFT targets under memory pressure.

\subsection{Schedulability Analysis}
%In this section, we discuss the boundary conditions of our scheduling system. 
% Our scheduling algorithm operates within well-defined capacity constraints, which is the rate of all requests in the working set does not exceed the system's current throughput $\Gamma$:  
Our scheduler enforces capacity constraints by ensuring the combined token generation rates within the working set $\mathcal{W}$ do not exceed the system's current throughput capacity $\Gamma$: 
\begin{equation}
    \sum_{i \in \mathcal{W}} r_i \leq \Gamma,
\end{equation}
where $r_i$ is request $i$'s generation rate. The throughput bound $\Gamma$ is dynamically estimated using real-time execution metrics, including request lengths and hardware utilization.
% where $\mathcal{W}$ denotes the working set and $r_i$ represents the required token generation rate of request $i$. The throughput capacity $\Gamma$ is dynamically estimated based on the execution time of currently running requests, input/output sequence lengths of active computations and hardware utilization metrics.

When this condition is violated (\ie $\sum r_i > \Gamma$), the system gracefully degrades to a \textit{first-come-first-served (FCFS)} policy with memory-aware admission control. In this fallback mode, requests are scheduled by arrival time while keeping the working set within available device memory. Excess requests remain offloaded to CPU memory until resources become available, and no new requests will be admitted.

%% file: text/5_memory_management.tex
\section{Hierarchical Memory Management} \label{sec:5-memory-management}
As the buffer-aware request scheduler optimizes heterogeneous request serving through request preemption, the KV cache management system must handle more concurrent requests than GPU memory can physically accommodate. This necessitates a hierarchical architecture supporting two fundamental operations: (a) \textbf{preempt} to offload KV cache from preempted requests, and (b) \textbf{resume} to reload cached data when requests regain execution. 

%\shengzhong{Who is the compared baseline in defining the following distinction?}
{Unlike popular LLM serving frameworks like SGLang, which manage KV cache reactively to optimize GPU throughput for dynamic batching, the key distinction of our approach lies in its proactive design philosophy. Existing systems typically adopt a reactive strategy, initiating data migration only after preemption decisions are made, which incurs substantial latency due to on-demand KV cache transfers. In contrast, TokenFlow exploits GPU I/O idle periods to preemptively offload KV caches to CPU memory. As a result, when preemption actually occurs, most of the cache has already been written back, rendering the KV cache offload operation nearly instantaneous.

To further align memory management with our proactive principle, we incorporate a write-through policy, synchronous chunked writing, and overlapped load-evict operations. Together, these techniques ensure efficient and concurrent data transfer, effectively narrowing the speed gap between computation and memory operations.}
%To support dynamic request preemption while minimizing KV cache transfer overhead, \method implements a hierarchical KV cache manager featuring: (1) write-through policy, (2) synchronous chunked writing, and (3) overlapped load-evict operations.

\subsection{Write-Through Policy}
We regard GPU memory as a high-speed cache for larger CPU memory, with two data write options: \textbf{write-through} and \textbf{write-back}. Traditional systems use a write-back policy, and KV cache writes are only triggered when the scheduler directs preemption. This approach suits static preemption patterns (\eg first-come-first-serve or multilevel feedback queues), allowing schedulers to proactively schedule writes and mask I/O latency during computation.

\begin{figure}[t!]
    \centering
    \includegraphics[width=0.9\linewidth]{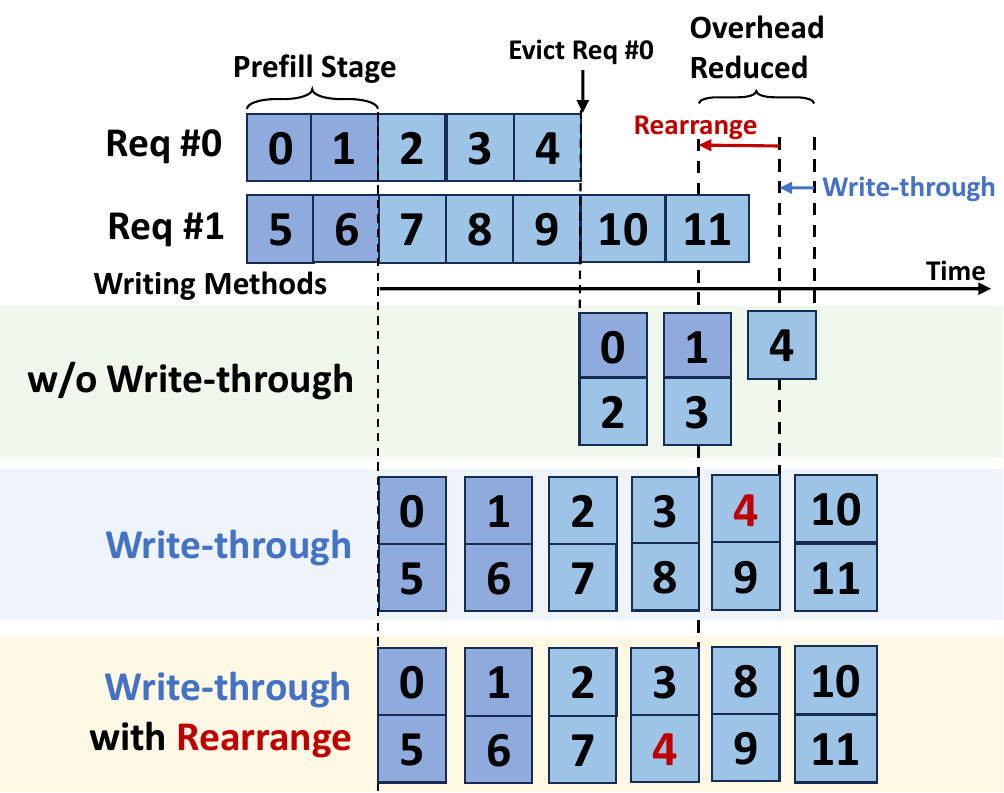}
    \caption{Comparison of Write Strategies: (Top) Conventional write-back approach (slowest); (Middle) Write-through method with reduced overhead (highlighted in blue); (Bottom) Rearranged write-through with additional optimizations (highlighted in red).}
    \label{fig:4-3-write-through}
\end{figure}

However, in our buffer-aware scheduler, requests may undergo multiple preemption-resumption cycles. Write-back policy becomes suboptimal due to unpredictable write timing and inability to pipeline I/O operations, leading to excessive I/O latency. Empirical measurements reveal that KV cache write speeds surpass amortized generation rates, motivating our adoption of a write-through policy. As shown in Figure~\ref{fig:4-3-write-through}, we continuously synchronize the generated KV cache to host memory, maintaining device-host cache consistency.
The write-through policy provides three key advantages: (1) eliminates the need for preemption pattern prediction, (2) fully utilizes PCIe write bandwidth between GPU-CPU memories, and (3) enables incremental updates – only newly generated KV cache entries require writing to host memory during resuming.

\begin{figure}[t!]
    \centering
    \includegraphics[width=0.95\linewidth]{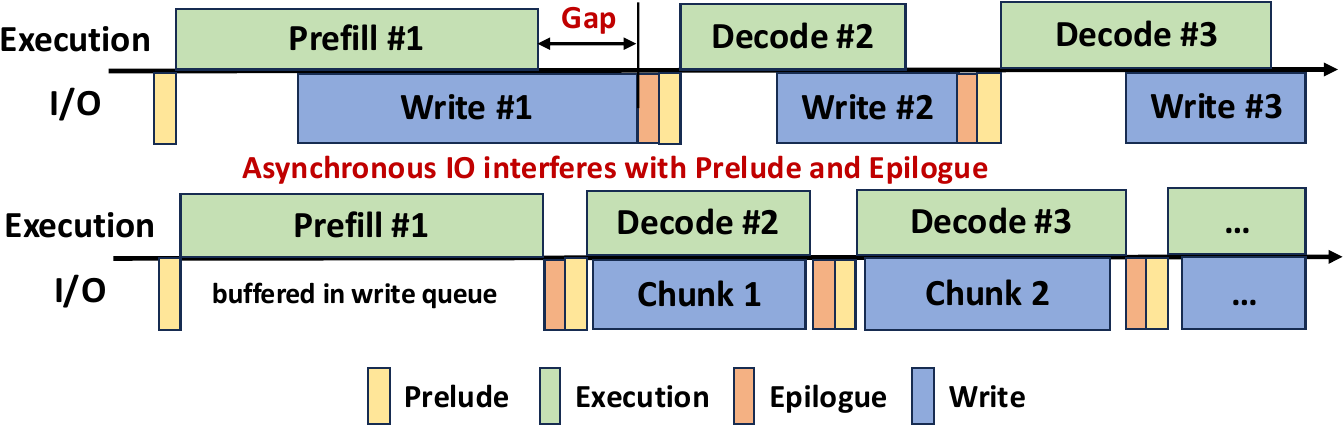}
    \caption{Temporal execution illustration of synchronous chunked writing scheme.}
    \label{fig:4-5-synchronize-write}
\end{figure}

\subsection{Synchronous Chunked Writing}
Effective offloading systems require overlapped I/O transfers, but traditional solutions prove inadequate in our case. Proactive transfers fail due to unpredictable preemption patterns, while asynchronous transfers create scheduling dependencies – iteration results must complete the transfer before subsequent scheduling decisions.

\method addresses this through a synchronous chunked writing scheme illustrated in Figure~\ref{fig:4-5-synchronize-write} that guarantees write operations complete within subsequent computation intervals. The mechanism operates as follows: (1) Each iteration buffers the generated KV cache;
(2) Pre-iteration phase: Scheduler estimates the next iteration's execution time;
(3) The offloading system pulls appropriate data chunks from the write buffer;
(4) Launches precisely sized write operations matching the estimated compute duration.

This approach achieves three benefits: (1) eliminates scheduler stalls from I/O completion waits, (2) maximizes PCIe bandwidth utilization through size-optimized transfers, and (3) enables priority-based write ordering. By analyzing scheduler request buffer sizes, \method prioritizes writes for requests with larger buffers (higher preemption probability), outperforming FIFO approaches as shown in Figure~\ref{fig:4-3-write-through}.

\subsection{Load-Evict Overlap}
\begin{figure}[t!]
    \centering
    \includegraphics[width=0.95\linewidth]{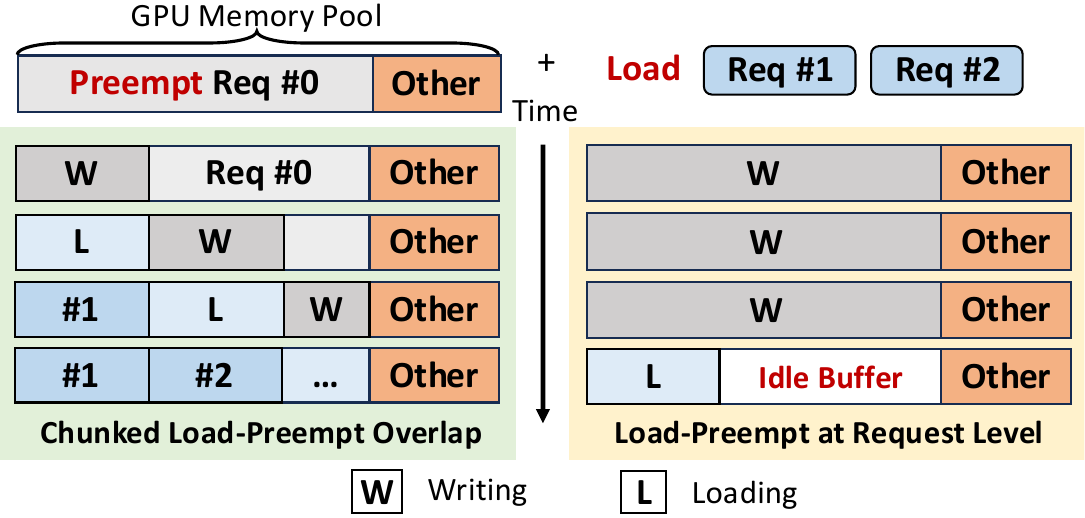}
    \caption{Load-Evict overlap technique in \method.}
    \label{fig:4-4-evict-load-overlap}
\end{figure}

Concurrent preemption and resume operations introduce challenges in transfer latency and memory contention. Baseline approaches trade-off between memory buffering (reducing latency) and operation serialization (minimizing memory), but \method resolves this through load-evict overlap.
The write-through policy enables partial memory reclamation during preemption – already synchronized KV cache segments can be immediately evicted. As visualized in Figure~\ref{fig:4-4-evict-load-overlap}, when preempting Request 0 while resuming Requests 1\&2: (1) Eviction of the remaining Request 0 KV cache proceeds concurrently. (2) Chunked loading of Requests 1\&2 KV cache overlaps with partial evictions. (3) Memory buffers are dynamically repartitioned during transfer.

Such overlapping reduces both transfer latency and memory fragmentation. Chunked writing allows precise control over transfer sizes, ensuring load-evict operations are complete within scheduler-determined time windows while maintaining memory safety guarantees.

%% file: text/6_implementation.tex
\section{Implementation}\label{sec:4_implementation}

We implement \method on top of SGLang, a high-performance Python framework for LLM serving, chosen for its modular architecture and efficient execution backend.
%While our implementation maintains full compatibility with existing SGLang applications, we emphasize that our core algorithmic contributions, including the preemptive scheduling policy and quality-of-service guarantees, represent general design principles that could be integrated into other serving frameworks with comparable KV cache management capabilities. Our KV cache manager is also not difficult to be ported to other LLM serving frameworks.
% 
% The system comprises a priority-based scheduler that maintains virtual buffers for each request to enforce rate guarantees and a hierarchical KV cache manager that handles hardware-level resource monitoring and allocation, which are implemented in approximately 4000 lines of Python code.
The system comprises a priority-based scheduler and a hierarchical KV cache manager, both of which are implemented with $\sim$4000 lines of Python code.

\myparacolon{Scheduler} 
%我们的调度算法理论上可以与任何一个推理框架的推理后端相适配，只需要这个推理后端实现了简单的preemption功能，但是想要完全发挥我们的调度算法的能力，就需要与我们设计的Hierarchical KV manager一起使用。利用SGLang的模块化特征，我们在SGLang的基础上完全重写了新的Scheduler替换原本的SGLang scheduler模块。作为baseline，我们还在SGLang上实现了基于recompute preemption的Andes算法。我们通过在SGLang中为每个请求维护一个虚拟的buffer来是的调度满足每个请求的速度需求，其余调度所需的硬件相关的数据都由KV Cache Manager负责收集。
Our scheduling algorithm works with any inference framework that supports basic preemption but performs best when integrated with our specialized Hierarchical KV Manager. By leveraging SGLang's modular architecture, we replaced its default scheduler with our optimized version and introduced request tracking and management modules to monitor buffer sizes and ensure stable output rates for each request. For benchmarking, we also implemented the Andes in SGLang using a recompute-based preemption approach.

\myparacolon{KV Cache Manager}
% The KV Cache Manager implementation leverages CUDA streams and Python multithreading to achieve concurrent execution of LLM inference and I/O operations, ensuring non-blocking interaction between these critical components. By employing dedicated CUDA streams for compute and memory operations alongside a multi-threaded architecture with proper synchronization mechanisms, we effectively overlap the computational workload with necessary memory transfers. Furthermore, the system implements batched write operations that consolidate numerous small memory updates into larger, more efficient transactions, significantly improving I/O throughput while maintaining low latency. This approach optimizes both hardware utilization and end-to-end performance by minimizing idle cycles and maximizing parallelism between computation and data movement. It is also worth mentioning that our KV cache manager design is also general and not difficult to be ported to other LLM serving frameworks.
Our hierarchical KV cache manager leverages parallel CUDA streams and Python multithreading to achieve fully overlapped compute and memory operations. The implementation maintains dedicated streams for computation, loading, and eviction, synchronized to maximize PCIe bandwidth utilization. Through dynamic chunk sizing and batched transfers, we continuously synchronize the KV cache between host and device memory while matching computation intervals. A central control thread coordinates these operations using CUDA events, maintaining non-blocking execution that supports dynamic request preemption with minimal overhead.

%% file: text/7_experiment.tex
\section{Evaluation} \label{sec:5_experiment}
% We evaluate our method on a variety of models, hardware configurations, and request datasets with varying input and output length characteristics, and find the following:
% \begin{itemize}
%     \item Summarize the highlight experiment result here.
%     \item Summarize the highlight experiment result here.
%     \item Summarize the highlight experiment result here.
% \end{itemize}

\subsection{Experimental Setups}
\label{sec:setup}

\subsubsection{Hardware and Models.}
To comprehensively evaluate the effectiveness of our approach in various hardware configurations, we performed experiments on multiple GPU platforms, including NVIDIA RTX 4090, A6000, and H200. 
To demonstrate the versatility of our system, we test on models of varying scales and architectures, including Llama3-8B, Qwen2-7B, and Qwen2.5-32B. %This diverse selection enables thorough validation of our system's adaptability across different model sizes and structures.

\subsubsection{Datasets.}
\label{sec:datasets}
\begin{figure}[t!]
    \centering
    \includegraphics[width=0.9\linewidth]{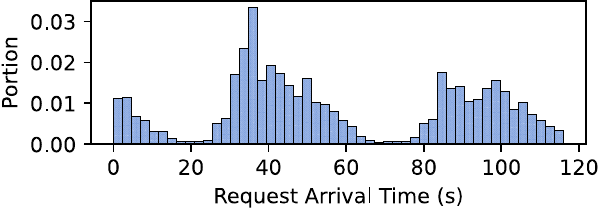}
    \caption{The distribution of the real-world trace.}
    \label{fig:6-realworld-trace}
\end{figure}

% To evaluate our system's end-to-end performance, we conduct comprehensive experiments using both public benchmarks and production traces. Our evaluation leverages two standard datasets: (1) ShareGPT, which provides a diverse collection of user prompts for assessing general performance, and (2) BurstGPT, a widely-adopted benchmark trace specifically designed for LLM service evaluation.
% % 
% Furthermore, to ensure practical relevance, we augment these standard benchmarks with real-world traces collected from production LLM services, whose distribution is shown in Figure~\ref{fig:6-realworld-trace}. These production traces capture authentic workload patterns and service characteristics. Additionally, we generate synthetic request distributions to systematically evaluate under various load conditions. 
To evaluate the end-to-end performance, we employ a combination of standard benchmarks and authentic production data. Our assessment utilizes two established datasets: ShareGPT~\cite{sharegpt2025} for general performance evaluation through diverse user prompts, and BurstGPT~\cite{wang2024burstgpt} as a specialized benchmark for LLM service analysis. Crucially, we incorporate real operational traces collected directly from production LLM services (distribution shown in Figure~\ref{fig:6-realworld-trace}), providing ground-truth workload patterns and service behaviors observed in actual deployments. Besides, we also use carefully constructed synthetic request distributions that systematically probe various load conditions.

\subsubsection{Evaluation metrics.}
%1. TTFT 2. TBT 3. RPS 4. Throughput 5. Valid throughput
To comprehensively assess the performance of our proposed serving system, we employ the following key metrics:
\begin{enumerate} [wide, labelwidth=!, labelindent=0pt, label=$\mathbf{\arabic*}.$]
    \item \textbf{TTFT (Time To First Token)}: Measures the latency from when a user submits a request to when the first token is generated. A lower TTFT indicates a more responsive system, crucial for user experience in interactive applications.
    \item \textbf{Throughput}: Refers to the total number of tokens generated per second. This metric reflects the system’s overall capacity to deliver output under a given workload.
    {
    \item \textbf{Effective Throughput}: Similar to throughput but inspired by video streaming experience, this metric evaluates real-time streaming performance by applying a timeliness-based weight to each token based on empirical observations of user consumption patterns:
    \begin{itemize}
        \item Tokens are fully counted when the buffer size is below $\tau_1 = 10\%$ of the total output length, as they are immediately consumable.
        \item Token contribution decays linearly to zero as the buffer grows from $\tau_1 = 10\%$ to $\tau_2 = 20\%$, reflecting tokens stored to mitigate network or resource fluctuations.
        \item Tokens beyond $\tau_2 = 20\%$ are excluded, as they exceed what is useful for a timely user experience.
    \end{itemize}
    This yields a more realistic measure of text-streaming Quality of Service (QoS) than conventional throughput.
    }
\end{enumerate}
% To demonstrate the superiority of our proposed serving system in streaming LLM outputs, we also introduce \textbf{effective throughput} as a key evaluation metric. Unlike conventional throughput measurements, effective throughput only accounts for tokens that successfully reach the user—whether delivered directly by the serving system or via the user buffer—while strictly adhering to the user-specified output speed. This metric holistically evaluates both system efficiency and compliance with Service-Level Objectives (SLOs), ensuring a balanced assessment of performance in real-world streaming scenarios.

\subsubsection{Baselines.}
% 1. conservative schedulers: SGLang default
% 2. QoS-aware schedulers: Andes
We evaluate our system against three categories of baseline schedulers:
\begin{enumerate}[wide, labelwidth=!, labelindent=0pt, label=$\mathbf{\arabic*}.$]
    \item \textbf{SGLang}~\cite{zheng2024sglang}: Conservative scheduling.
    \item \textbf{SGLang (chunked)}: SGLang with chunked prefill.
    \item \textbf{Andes}~\cite{liu2024andes}: QoE-aware scheduler for text-streaming SLAs.
\end{enumerate}

\begin{figure}[t!]
    \centering
    \includegraphics[width=1\linewidth]{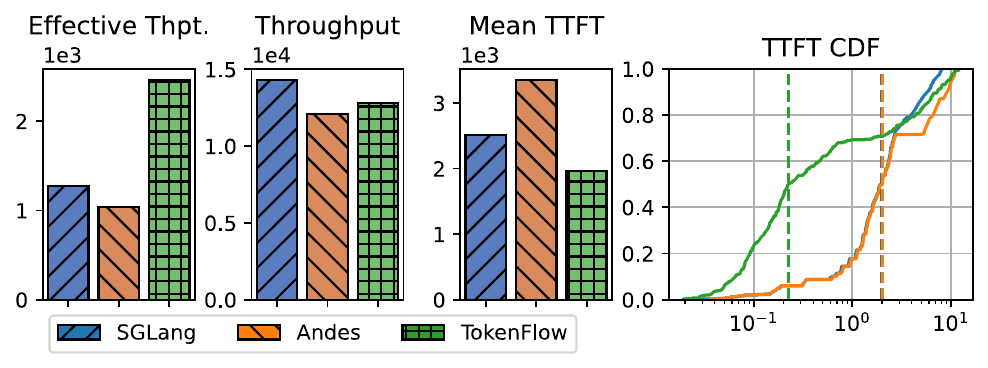}
    \caption{End-to-end performance metrics on H200 with Llama3-8B model.}
    \label{fig:6-e2e-H200}
\end{figure}

\begin{figure}[t!]
    \centering
    \includegraphics[width=1\linewidth]{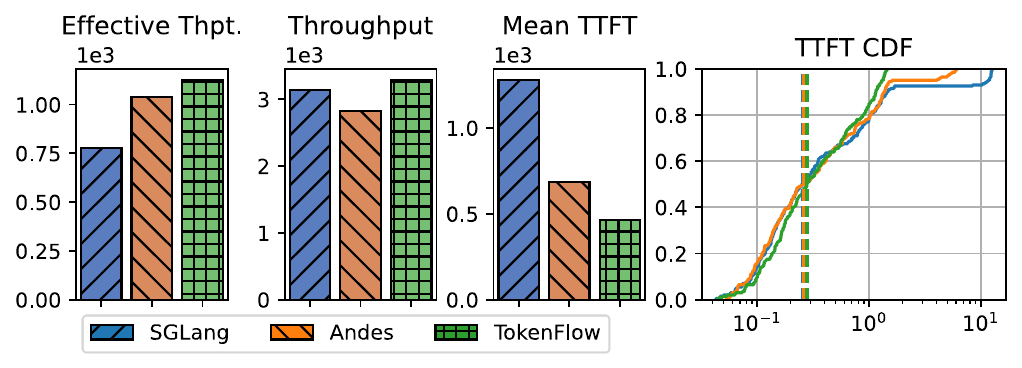}
    \caption{End-to-end performance metrics on A6000 with Qwen2.5-8B model.}
    \label{fig:6-e2e-a6000}
\end{figure}

\subsection{End-to-end Evaluation in Real-World Traces}
To validate \method's effectiveness under production conditions, we conduct large-scale experiments using two categories of request traces introduced in Section \ref{sec:datasets}. %: (1) public benchmarks (BurstGPT with ShareGPT prompts) for standardized comparison, and (2) proprietary traces collected from live LLM services to simulate real deployment scenarios. 
All experiments are performed on NVIDIA H200 and A6000, paired with two series of models, Llama3 and Qwen2.5. %, representing diverse hardware capabilities from data-center to edge server.

\mypara{Effective Throughput and TTFT Improvement}
Our experimental results demonstrate significant improvements across all evaluation metrics when comparing \method to baseline systems (SGLang and Andes) on both BurstGPT and proprietary traces. As shown in Figure~\ref{fig:6-e2e-H200}, Figure~\ref{fig:6-e2e-a6000}, \method achieves an average 52.6\% reduction in mean TTFT (up to 88.7\% at P50) while increasing effective throughput by 45.1\% on A6000 and 37.1\% on H200. 
% 
%The performance gains are particularly pronounced on H200 GPUs, where the combination of our buffer-aware scheduling and the hardware's high memory bandwidth delivers  better throughput scaling efficiency compared to SGLang under heavy load conditions. 
The results validate that \method's co-design of scheduling and memory management successfully translates theoretical advantages into measurable improvements across diverse hardware configurations and workload patterns.

% \begin{figure}[t!]
%     \centering
%     \includegraphics[width=1\linewidth]{figure/6/6-e2e-4090.pdf}
%     \caption{End-to-end performance metrics on RTX 4090: throughput, effective throughput, and TTFT distribution.}
%     \label{fig:6-e2e-4090}
% \end{figure}
\begin{figure}[t!]
    \centering
    \includegraphics[width=1\linewidth]{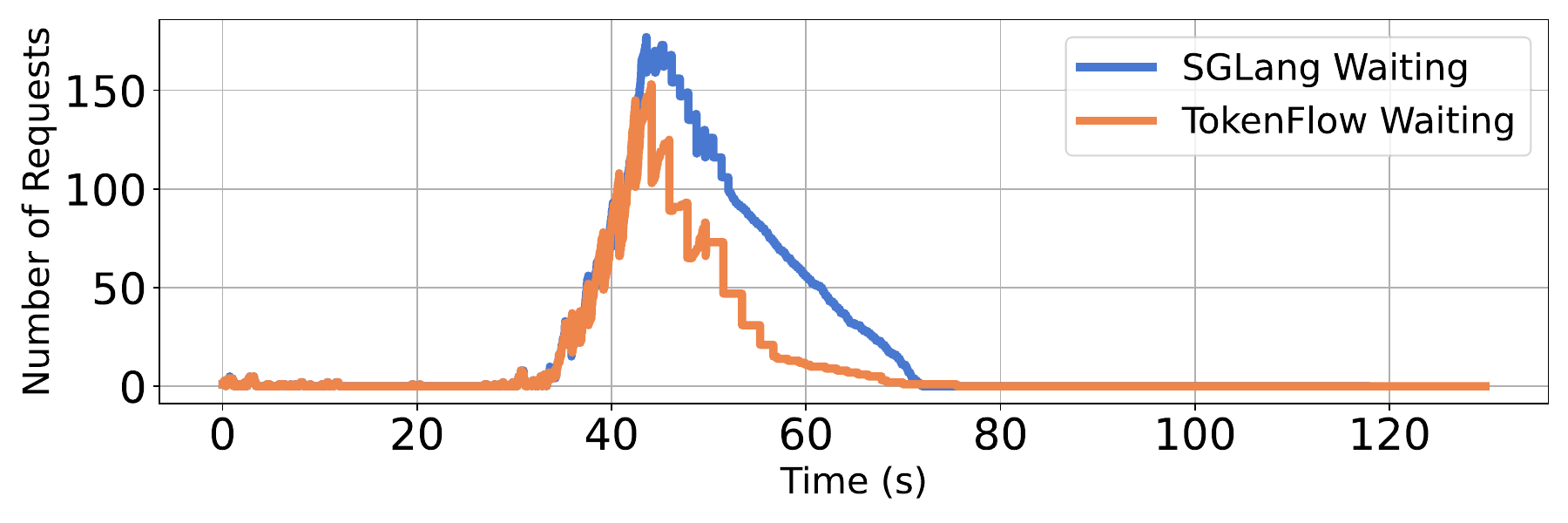}
    \caption{Temporal variation of \textbf{\underline{queued requests}} during a representative trace segment.}
    \label{fig:6-waiting-requests}
\end{figure}

\begin{figure}[t!]
    \centering
    \includegraphics[width=1\linewidth]{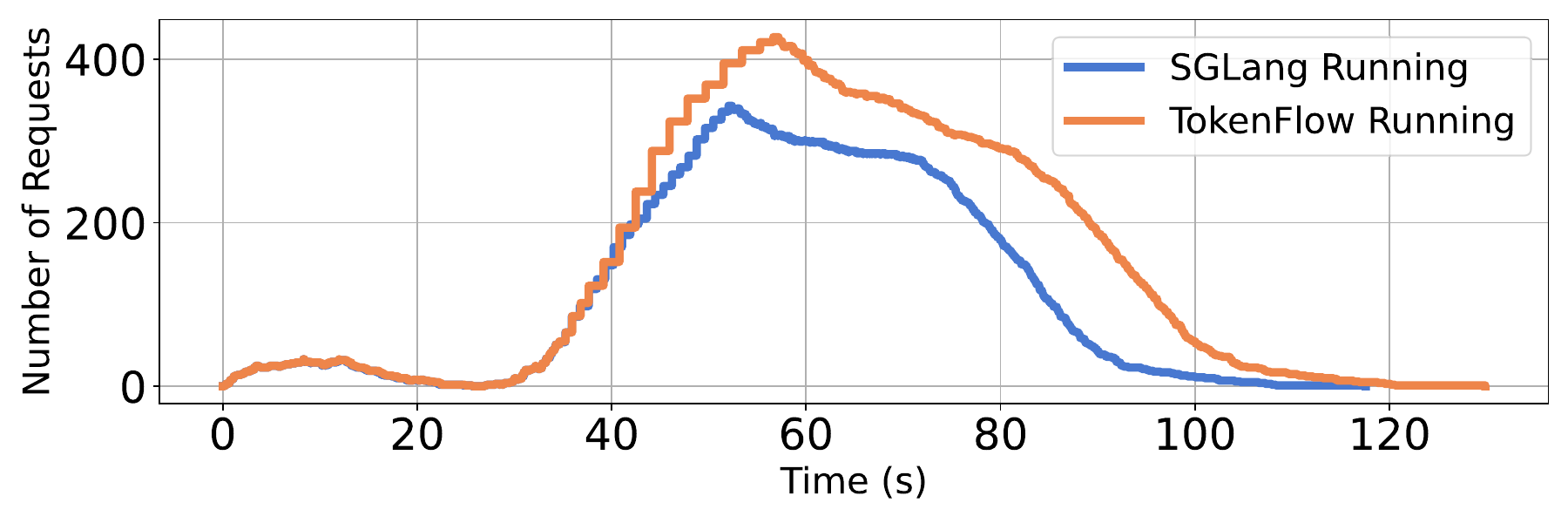}
    \caption{Temporal variation of \textbf{\underline{running requests}} during a representative trace segment. }
    \label{fig:6-running-requests}
\end{figure}

\mypara{Experimental Validation with Long-term Trace}
% 我们还尝试在H200上服务Qwen2.5 32B模型，重放BurstGPT约20分钟的trace，并监控系统实时的等待请求数和运行请求数。我们截取了一部分等待请求数和运行请求数的结果，分别如图Figure~\ref{}和Figure~\ref{}所示。我们发现在服务繁忙的场景下\method的等待请求数要明显少于Baseline，同时并发运行的请求数也要明显多于Basline，这主要是因为我们设计的调度策略和高效的内存管理是的系统可以超额服务请求。
To evaluate system scalability, we stress-tested the Qwen2.5-32B model on H200 GPUs using a 20-minute BurustGPT trace while monitoring real-time metrics. As shown in Figures~\ref{fig:6-waiting-requests} and~\ref{fig:6-running-requests}, \method outperforms baselines under peak load, with fewer queued requests and higher concurrency. This is enabled by its coordinated design: a buffer-aware scheduler for dynamic prioritization and hierarchical memory management for efficient KV cache swapping.

\subsection{Controlled Request Distribution Test}
\label{sec:controlled-test}
We evaluate \method’s performance using synthetic workloads that reflect real-world patterns from ShareGPT~\cite{sharegpt2025}, enabling systematic comparison with baseline methods in terms of throughput and latency. Experiments are conducted on NVIDIA H200 (starting with \texttt{mem-frac=0.3}) and RTX 4090 GPUs under two scenarios: (1) Bursty arrivals (burst size $b$) simulating flash crowds and (2) Poisson-distributed (rate $\lambda$) arrivals modeling typical traffic. Input/output lengths follow normal distributions, with the RTX 4090 using 512/1024-token inputs in average (S/L) and 1024/2048-token outputs in average (S/L), while H200 outputs are scaled 2$\times$. (Here, S denotes ``Short'' and L denotes ``Long'' sequence lengths.) Full configurations are provided in Table~\ref{tab:6-3-experiment-setup}.

\input{table/6/6-3-experiment-setup}

\begin{figure}[t!]
    \centering
    \includegraphics[width=1\linewidth]{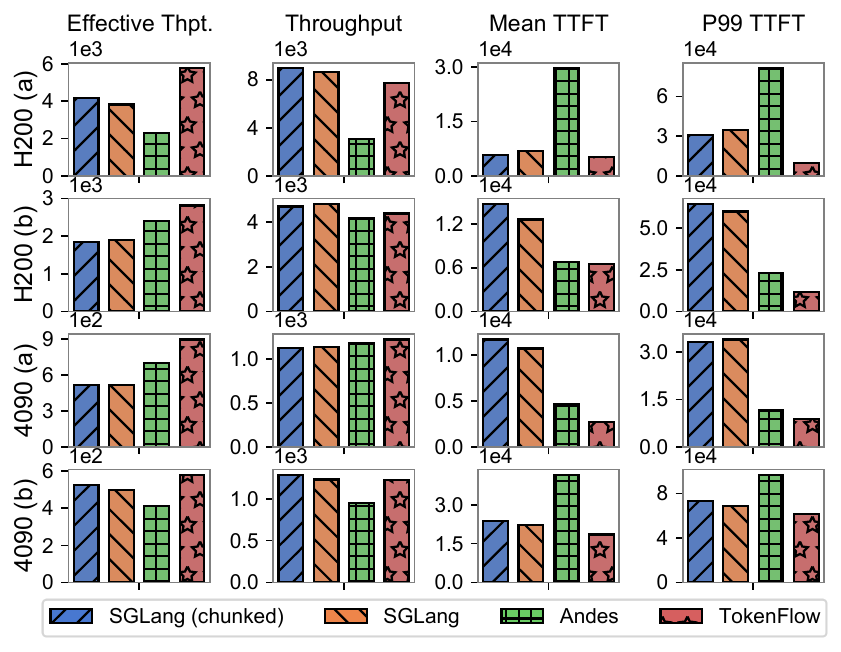}
    \caption{Performance Metrics During Burst Workload. }%\shengzhong{Which GPU model?}}
    \label{fig:3-burst}
\end{figure}

\mypara{Improvement under burst request scenarios}
We present effective throughput, raw throughput, mean TTFT, and P99 TTFT across various experiment setups in Figure~\ref{fig:3-burst}. \method outperforms Andes and SGLang across all metrics, achieving: (1) up to 80.2\% lower P99 TTFT, (2) up to 48.4\% lower mean TTFT, and (3) up to 52.9\% higher effective throughput. While Andes shows notable degradation compared to SGLang in throughput, \method retains similar computational efficiency to SGLang.

These performance gains stem from architectural differences. SGLang’s rigid FCFS and prefill-prioritized scheduling struggle with dynamic workloads. Andes improves QoS via quality-aware scheduling but neglects throughput efficiency. In contrast, \method’s co-design balances user-perceived latency and system throughput through intelligent scheduling and buffer-aware resource management.

\begin{figure}[t!]
    \centering
    \includegraphics[width=1.00\linewidth]{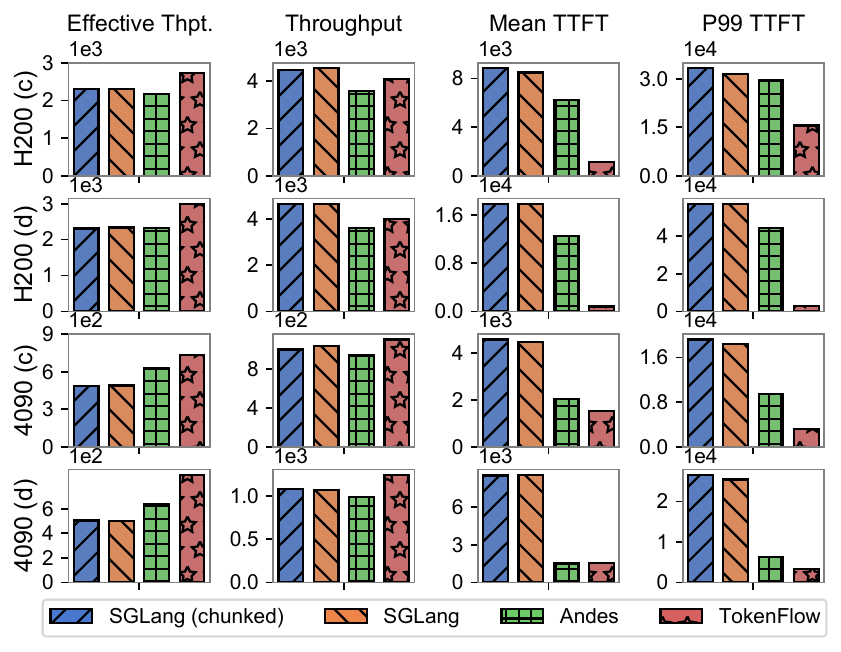}
    \caption{Performance Metrics During Poisson Workload.}
    \label{fig:3-poisson}
\end{figure}

\mypara{Improvement under Poisson-distributed request scenarios}
Figure~\ref{fig:3-poisson} quantifies system performance under Poisson workloads on four key metrics: effective throughput, raw throughput, mean TTFT, and P99 TTFT. \method outperforms baselines in all scenarios, notably improving effective throughput by 82.5\% (RTX 4090) and reducing TTFT by 53.7\% (H200) versus the competitors.
Under heavy load, when GPU memory becomes saturated, SGLang suffers from severe queuing delays that drastically increase TTFT, while Andes fails to maintain balanced streaming performance. \method overcomes these limitations and delivers both low latency and consistent output rates even at peak loads. 

\begin{figure}
    \centering
    \includegraphics[width=1\linewidth]{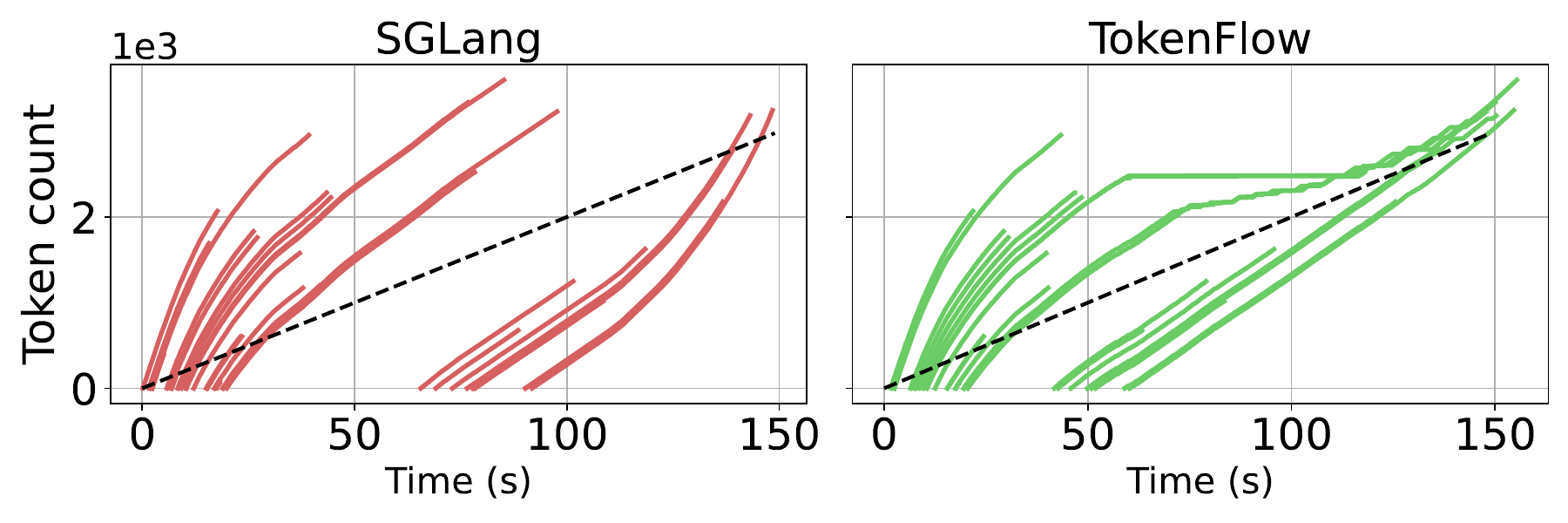}
    \caption{Token generation timelines comparing SGLang (left) and \method (right). \method maintains generation speeds consistently above requirements (black line) while achieving lower average TTFT.} %\shengzhong{shorter, thicker, grid}}
    \label{fig:6-token-generation-timeline}
\end{figure}

\subsection{Micro Experiments}
\myparacolon{$\blacktriangleright$Token Generation Timeline} 
Our qualitative comparison with SGLang’s scheduling in Figure~\ref{fig:6-token-generation-timeline} highlights \method’s dual advantages: it initiates service earlier while maintaining precise token delivery at required speeds, unlike SGLang’s head-of-line blocking, which forces subsequent requests into prolonged waiting. The timeline analysis demonstrates how \method’s preemptive scheduling leverages per-request buffers to ensure stable throughput and timely service, generating each token at optimal consumption rates.
%Our qualitative comparison with traditional SGLang scheduling, illustrated in Figure~\ref{fig:6-token-generation-timeline}, reveals \method's dual advantages through token generation timeline analysis: \method not only initiates service earlier than SGLang's rigid execution strategy but also maintains perfectly aligned token delivery at required speeds, while SGLang's head-of-line blocking forces subsequent requests to endure prolonged waiting periods. The visual comparison demonstrates how \method's preemptive scheduling leverages per-request buffers to achieve stable throughput and timely service, consistently generating each token at ideal consumption speeds.

\myparacolon{$\blacktriangleright$Visualize Preemptive Scheduling} 
Figure~\ref{fig:6-token-generation-timeline} shows \method's preemption mechanism: when a request’s token buffer reaches a threshold, resources are reallocated to improve throughput. Preempted requests pause (seen as plateaus in timelines) without harming latency, resuming only when buffers near depletion. This ensures guaranteed QoS across concurrent requests.

\begin{figure}[t!]
    \centering
    \includegraphics[width=0.9\linewidth]{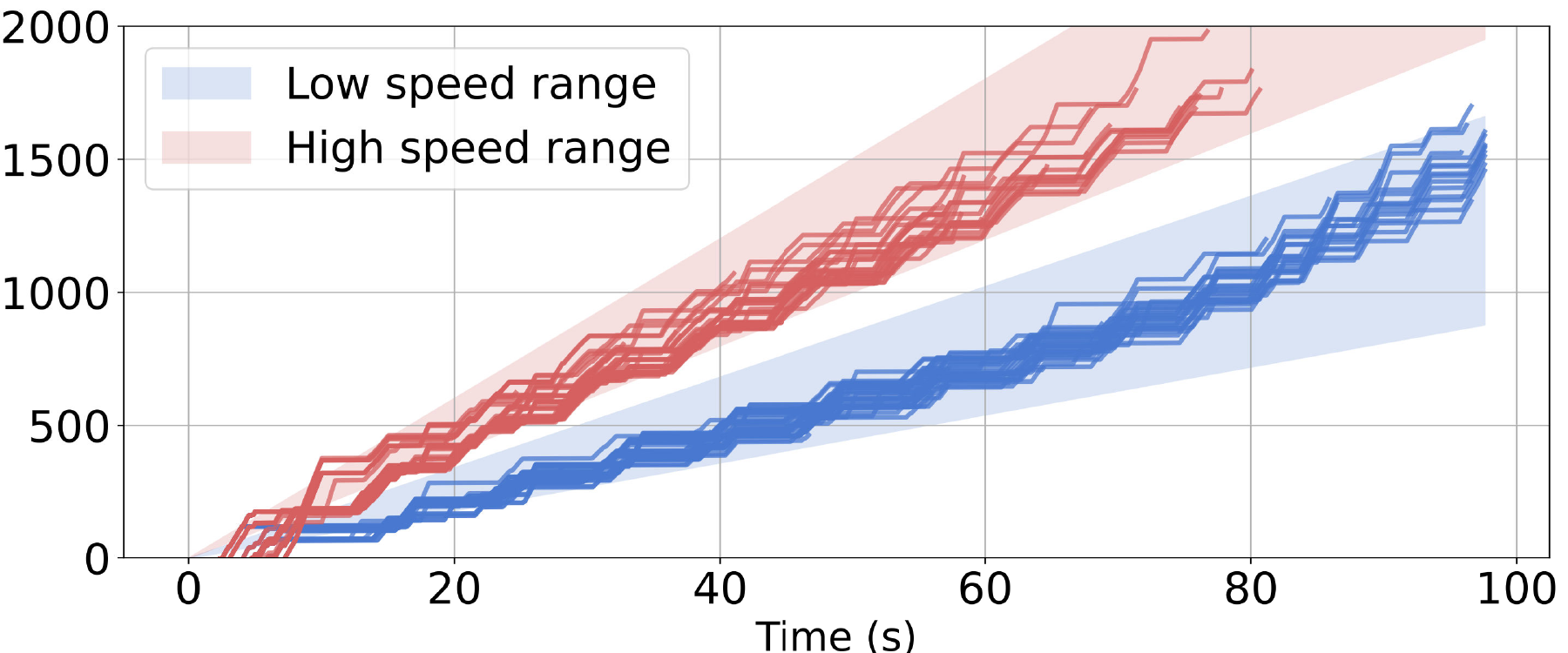}
    \caption{Experiment on multi-rate request scheduling.}
    \label{fig:6-different-output-speed}
\end{figure}

\myparacolon{$\blacktriangleright$Multi-Rate Request Scheduling} 
In evaluating \method with a mixed-rate burst workload (40\% at 15 tokens/s, 60\% at 20 tokens/s), Figure~\ref{fig:6-different-output-speed} shows distinct timelines where each request type consistently maintains its target rate within tolerance bands. This automatic rate differentiation emerges from \method's buffer-aware prioritization - higher-rate requests naturally drain buffers faster, gaining implicit scheduling priority. The system thus supports heterogeneous rates while maintaining strict QoS guarantees without manual configuration.
% We evaluated \method under a mixed-rate burst workload: 40\% requiring 15 tokens/sec (low-rate) and 60\% demanding 20 tokens/sec (high-rate). As visualized in Figure~\ref{fig:6-different-output-speed}, the token generation timelines distinctly segregate into blue (15 tokens/sec) and red (20 tokens/sec) bands, each consistently maintaining their target rates within the shaded tolerance intervals. This emergent behavior stems from \method's inherent buffer-aware prioritization, requests with higher rate naturally drain buffers quicker, gaining implicit scheduling priority. This buffer-aware prioritization enables \method to support heterogeneous rates while preserving strict QoS guarantees—without manual configuration.

\begin{figure}[t!]
    \centering
    \includegraphics[width=\linewidth]{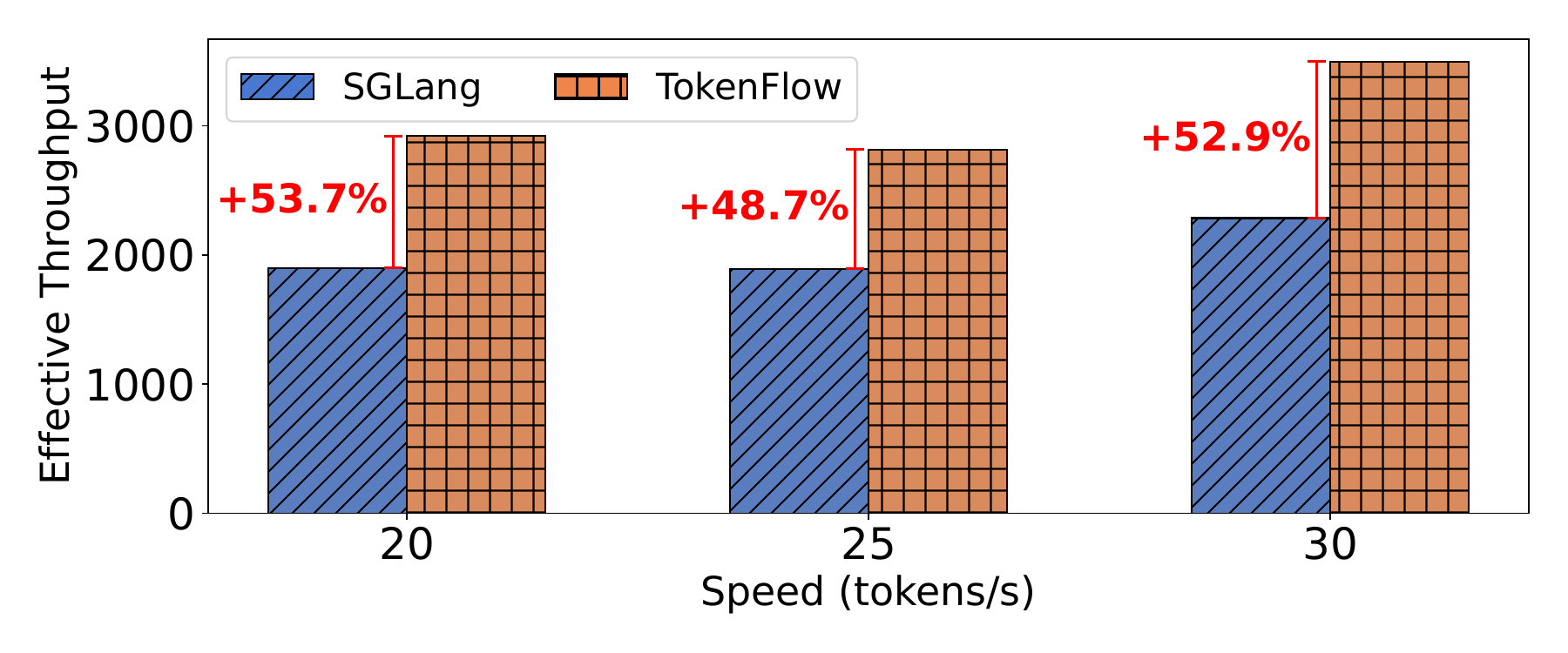}
    \caption{Effective throughput gains over different generation speeds.}
    \label{fig:6-improve-over-varying-rate}
\end{figure}
\myparacolon{$\blacktriangleright$Performance Across Diverse Generation Speed} We test workloads with requests at 20, 25, and 30 tokens/s to assess how TokenFlow adapts to varying generation speeds. As shown in Figure~\ref{fig:6-improve-over-varying-rate}, TokenFlow consistently achieves higher throughput and lower latency than baseline across all speed settings. 

\begin{figure}[t!]
    \centering
    \includegraphics[width=\linewidth]{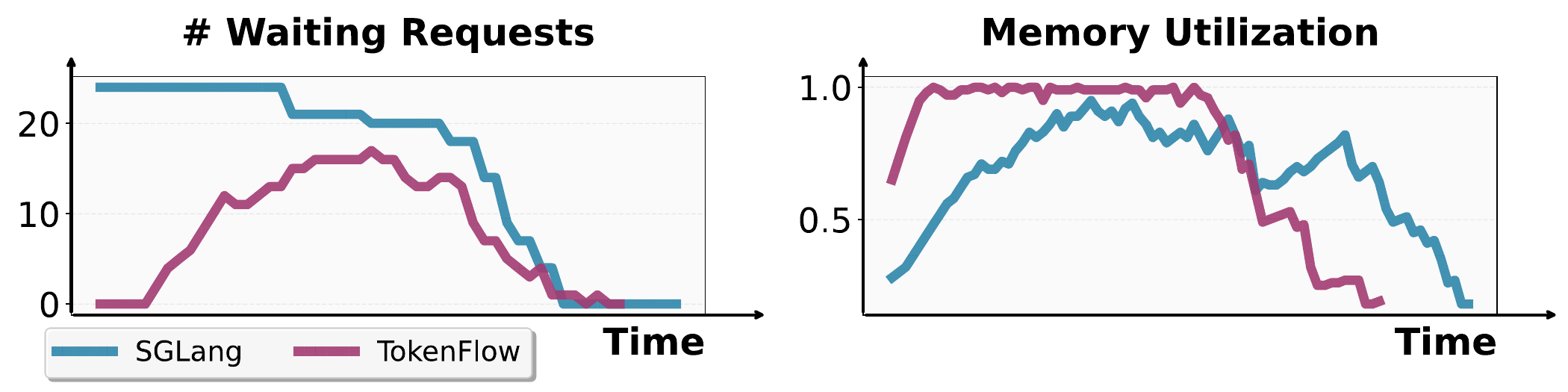}
    \caption{Performance on Huawei Ascend 910B}
    \label{fig:ascend-queue-req-and-memory-usage}
\end{figure}

\myparacolon{$\blacktriangleright$Diverse Hardware Support}TokenFlow scales efficiently across hardware, including Huawei Ascend 910B, maintaining high performance under bursty workloads (Figure~\ref{fig:ascend-queue-req-and-memory-usage}).

\subsection{Hyperparameter Sensitivity}
\begin{figure} [t!]
    \centering
    \includegraphics[width=1\linewidth]{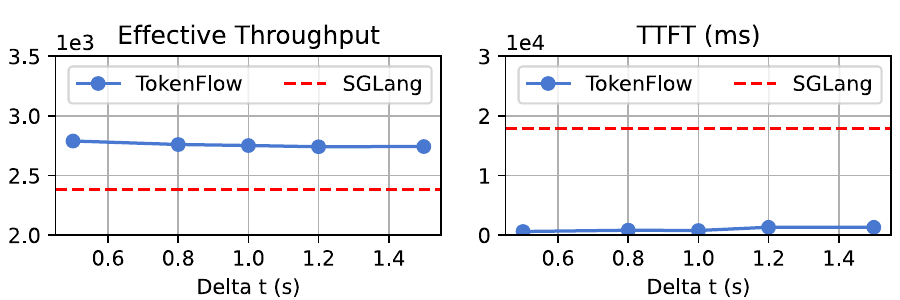}
    \caption{Impact of Rescheduling Interval $\Delta t$ on TTFT and Effective Throughput.}
    \label{fig:6-reschedule-interval-sensitivity}
\end{figure}

% 1. reschedule interval
% 2. buffer conservativeness
\mypara{Reschedule Interval}
Our study examines how rescheduling interval length ($\Delta t$) impacts system performance through its effects on buffer awareness and scheduling efficiency. As depicted in Figure~\ref{fig:6-reschedule-interval-sensitivity}, varying $\Delta t$ between 0.5-1.5 seconds reveals that shorter intervals marginally improve effective throughput and TTFT. While frequent updates enable better adaptation to dynamic conditions, they also incur scheduling overhead. The optimal interval should therefore balance responsiveness with computational cost, tailored to specific workload patterns and QoS requirements.

\begin{figure} [t!]
    \centering
    \includegraphics[width=\linewidth]{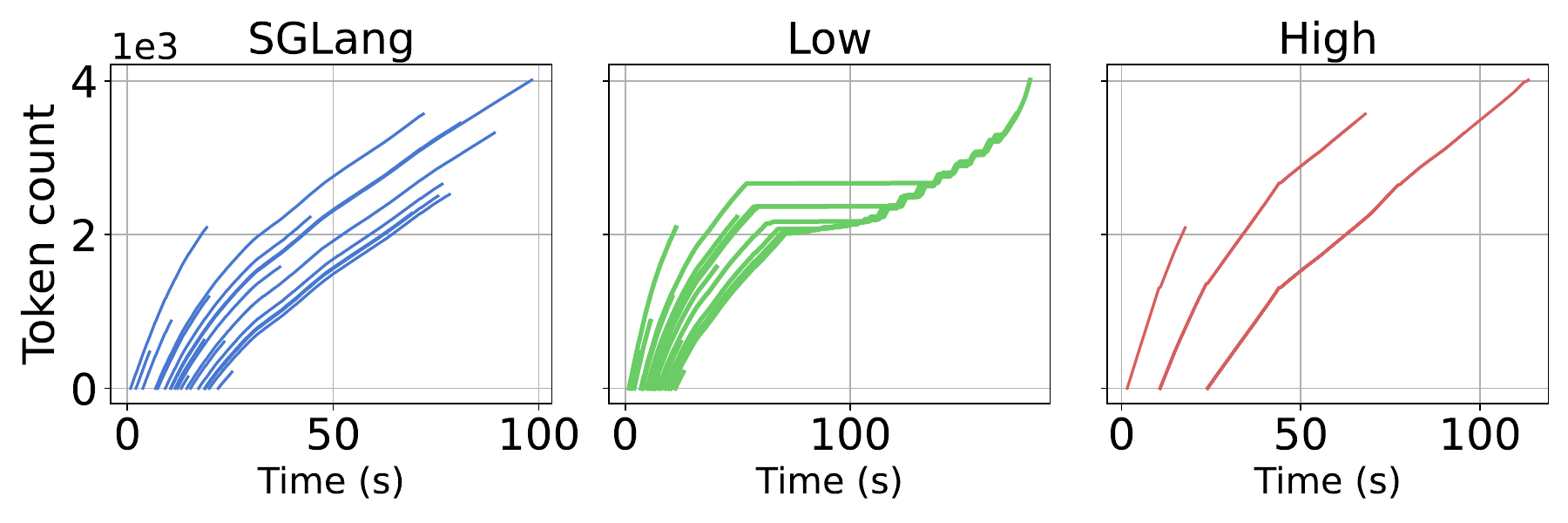}
    \caption{Impact of Buffer Conservativeness on Scheduler Behavior. Left: SGLang. Middle and Right: \method with different buffer conservativeness settings.}
    \label{fig:6-buffer-conservativeness-sensitivity}
\end{figure}

\mypara{Buffer Conservativeness} 
Our buffer conservativeness parameter controls how aggressively resources are reallocated based on request buffer levels. Experiments with high (20.0) and low (1.0) settings (Figure~\ref{fig:6-buffer-conservativeness-sensitivity}) reveal its critical role in balancing responsiveness and stability: higher values produce cautious, SGLang-like behavior favoring stability, while lower values enable agile workload adaptation at potential stuttering risk. This tunable parameter offers precise control over the responsiveness-stability tradeoff, allowing customization for diverse QoS needs.
% Our buffer conservativeness parameter governs how aggressively the scheduler reallocates resources based on request buffer levels. Through experiments with high (20.0) and low (1.0) conservativeness settings (Figure~\ref{fig:6-buffer-conservativeness-sensitivity}), we demonstrate this parameter's critical role in balancing system responsiveness and stability. Higher values (right) yield cautious, SGLang-like behavior that prioritizes stability, only rebalancing when buffers grow large. Lower values (middle) enable agile adaptation to workload changes at the risk of potential stuttering. This tunable parameter provides precise control over the responsiveness-stability trade-off, allowing customization for diverse QoS requirements, a key advantage of our system.

\subsection{Overhead Quantification and Ablation study}
{
\mypara{Overhead Analysis} We evaluate the scheduling overhead introduced by our scheduling algorithm and the newly designed request manager. For the scheduling algorithm, the runtime cost increases only marginally, from the negligible $\sim$0.07 ms of SGLang to a still minimal $\sim$0.4 ms. The Request Tracker and Manager, which handle metadata maintenance and priority queue management, contribute negligible overhead relative to the KV cache manager’s I/O cycles and the model’s forward computation.}

\mypara{Ablation Study} We also performed an ablation study to evaluate the individual contributions of each hierarchical memory management submodule, which uses the setting 4090 (b) as described in Section~\ref{sec:controlled-test}. Quantitative results are presented in Table~\ref{tab:6-6-ablation}. As shown in Table~\ref{tab:6-6-ablation}, \method's unique write-through and hierarchical offload design provide the most significant performance gains, demonstrating their critical role in our system's overall improvement.
% \input{table/6/6-6-overhead}
\input{table/6/6-6-ablation}

% \begin{table}[t]
% \caption{Performance of backbone models.}
% \label{tb:backbone}
% \small
% \begin{tabular}{c|cc|cc|cc}
% \toprule
% \textbf{Datasets} & \multicolumn{2}{c|}{\textbf{RealWorld-HAR}} & \multicolumn{2}{c|}{\textbf{WESAD}} & \multicolumn{2}{c}{\textbf{MOD}} \\ 
% \textbf{Metric (\%)} & \textbf{Acc.} & \textbf{F1} & \textbf{Acc.} & \textbf{F1} & \textbf{Acc.} & \textbf{F1} \\ \midrule
% \textbf{DeepSense} & 86.09 & 86.71 & 80.98 & 75.98 & 87.27 & 87.16 \\
% \textbf{ResNet} & 88.45 & 89.23 & 80.09 & 79.81 & 83.66 & 83.48 \\
% \textbf{Transformer} & 81.99 & 81.35 & 79.34 & 78.88 & 62.16 & 61.43 \\
% \bottomrule
% \end{tabular}
% \end{table}

%% file: table/6/6-3-experiment-setup.tex
\begin{table}[t!]
\centering
\caption{Experimental Configurations for Controlled Request Distributions Evaluation.}
\label{tab:6-3-experiment-setup}
\resizebox{0.75\linewidth}{!}{
\begin{tabular}{ccc}
\toprule
\textbf{Setup} & \textbf{RTX 4090} & \textbf{H200} \\  % \textbf加粗
\midrule
(a) & Burst $b=60$, SL & Burst $b=400$, SL \\
(b) & Burst $b=80$, LL & Burst $b=200$, LL \\
(c) & Poisson $\lambda=2$, SL & Poisson $\lambda=5$, SL \\
(d) & Poisson $\lambda=4$, SL & Poisson $\lambda=10$, SL \\
\bottomrule
\end{tabular}
}
\end{table}

%% file: table/6/6-6-ablation.tex
% \usepackage{booktabs}
\begin{table}[t!]
\centering
\caption{Ablation study of \method.}
\label{tab:6-6-ablation}
\resizebox{0.85\linewidth}{!}{
\begin{tabular}{@{}cccc@{}}
\toprule
\method & \small{\begin{tabular}[c]{@{}c@{}}w/o. \\ Offload\end{tabular}} & \small{\begin{tabular}[c]{@{}c@{}}w/o. \\ Write-Through\end{tabular}} & \small{\begin{tabular}[c]{@{}c@{}}w/o. \\ Evict-Load Overlap\end{tabular}} \\ \midrule
66.00 s & 127.28 s & 82.76 s & 74.43 s \\ \bottomrule
\end{tabular}
}
\end{table}

%% file: text/discussion.tex
\section{Discussion} \label{sec:discussion}
\mypara{Design Advantages over Andes} While Andes primarily emphasizes user-perceived Quality of Experience (QoE), it overlooks the broader perspective of overall inference server performance and resource utilization. In contrast, our system introduces a comprehensive Quality of Service (QoS) metric that unifies latency, throughput, and user experience, thereby offering a more holistic characterization of streaming quality. Furthermore, TokenFlow enhances preemption efficiency through a hierarchical memory manager that tightly cooperates with a two-stage scheduler. The scheduler conveys critical preemption requirements to the memory manager, while the memory manager, in turn, supplies feedback to refine scheduling decisions. This bidirectional interaction enables more effective resource allocation than Andes, which lacks such coordinated mechanisms.

\mypara{Scaling TokenFlow for Multi-Node and Distributed Systems}
Mooncake~\cite{305212} is a KV-centric LLM serving system that adopts a disaggregated architecture, leveraging RDMA to manage a distributed, multi-layer KV cache across nodes with a focus on throughput at scale. In contrast, TokenFlow follows a different design philosophy: rather than prioritizing distributed throughput, we emphasize single-node efficiency. TokenFlow employs proactive PCIe-based transfers to optimize the local GPU–CPU memory hierarchy and improve preemption. Although Mooncake is built for cross-node communication, TokenFlow’s scheduling and KV management can be extended to multi-node environments by introducing an inter-node cache layer and leveraging our co-designed scheduler to ensure KV consistency across nodes. Furthermore, TokenFlow mitigates PCIe bandwidth contention by adaptively reducing preemption frequency, complementing Mooncake’s RDMA-based architecture with optimized local data flow to provide a more holistic solution.

\mypara{Handles Different Client Types} TokenFlow requires user-facing clients to explicitly specify their desired output rate, which the server then enforces to ensure smooth streaming. For non-user consumers (e.g., LLM agents), we instead employ a reference rate as an indicator of scheduling priority: a larger reference rate signals higher priority, while a smaller value denotes lower priority. In future work, we plan to relax this requirement by allowing the scheduler to infer effective rates from request-level signals such as system load. For example, non-user requests may start at a low rate and accelerate when resources permit, then be throttled again under heavy load. Such adaptive control removes the need for explicit parameters while improving fairness and efficiency across heterogeneous clients.

%% file: text/8_related.tex
\section{Related Work} \label{sec:6_related}

% \paragraph{LLM Serving and Scheduling.}
% Recent LLM serving systems aim to improve throughput and latency via batching, memory optimization, as well as scheduling prefill and decode to match their computational characteristics~\cite{zheng2023response,280922,kwon2023efficient,zheng2024sglang,agrawal2024taming,patel2024splitwise,zhong2024distserve,wu2024loongserve,miao2024spotserve,ong2025routellm,griggs2024m}. However, these works underexplore request concurrency and dynamic scheduling. Traditional schedulers such as those in vLLM and SGLang rely on FCFS and static memory allocation, which falter under streaming or dynamic load. More advanced approaches like slice-level control, Past-Future pipelining, live migration, and QoE-aware pacing attempt to address these limitations but still lack full adaptability to streaming contexts and memory pressure~\cite{kossmann2024gpu,jin2024p,kwon2023efficient,zheng2024sglang,cheng2024slice,10.1145/3715014.3722067,10.5555/3691938.3691948,liu2024andes,du2024sida,10.1145/3676641.3716011}. 
\mypara{LLM serving systems}
Orca~\cite{280922} batches requests at the iteration level, building on the auto-regressive generation pattern of LLMs. vLLM~\cite{kwon2023efficient} improves LLM throughput by optimizing memory utilization. SGLang~\cite{zheng2024sglang} co-designs the front-end language and the back-end runtime for efficiency. Sarahti-Serve~\cite{agrawal2024taming} introduces a fragmented prefill to reduce the spike latency caused by large initial requests. Splitwise~\cite{patel2024splitwise}, DistServe~\cite{zhong2024distserve}, and LoongServe~\cite{wu2024loongserve} disaggregate the prefill and decode phases based on their different computation patterns. However, these LLM serving systems overlook the potential to improve request concurrency. Recently, other work has explored different aspects of LLM service: SLOs-Serve~\cite{chen2025slosserveoptimizedservingmultislo} optimizes multi-SLO service scenarios, and VoltanaLLM~\cite{yu2025voltanallmfeedbackdrivenfrequencycontrol} focuses on energy-efficient LLM service via feedback-driven frequency control.

\mypara{LLM request scheduling}
vLLM~\cite{kwon2023efficient} and SGLang~\cite{zheng2024sglang} adopt FCFS policies and allocate memory in advance. Prefill-first scheduling is simple but inefficient under dynamic loads. Slice-level scheduling~\cite{cheng2024slice} offers control over the granularity of scheduling. LightLLM~\cite{10.1145/3676641.3716011} uses a Past-Future scheduler to pipeline prefill and generation. Llumnix~\cite{10.5555/3691938.3691948} implements request scheduling using live migration across LLM instances. However, these schedulers are not designed for streamed content generation. Andes~\cite{liu2024andes} proposes QoE-aware scheduling using a Token Pacer to optimize perceived latency. However, it fails to adapt to buffer size or memory pressure during streaming. More advanced solutions explore hardware-level arbitration, disaggregated prefill/decode clusters, and MoE-aware scheduling~\cite{kossmann2024gpu,jin2024p,du2024sida,zhu2025nanoflowoptimallargelanguage}, but streaming remains an afterthought.

\mypara{Memory management for LLM serving}
It is the current standard to use KV cache~\cite{pope2023efficiently} in decoding phase. PagedAttention~\cite{kwon2023efficient}, RadixAttention~\cite{zheng2024sglang}, ChunkAttention~\cite{ye2024chunkattention} optimize GPU memory utilization by memory paging and sharing common prefixes between requests. CachedAttention~\cite{298501}, FastServe~\cite{wu2023fast} and Pensieve~\cite{yu2025stateful} develop hierarchical KV cache management for multi-turn conversation and request preemption. FlexGen~\cite{sheng2023flexgen} DeepSpeed Inference~\cite{aminabadi2022deepspeed} and SpInfer~\cite{10.1145/3689031.3717481} offload model weights. Request resumption can be accelerated by pipelining and blending multiple pre-computed KV cache chunks~\cite{10.1145/3689031.3696098,gim2024prompt,gao2025fast}. Lina~\cite{li2023accelerating}, PowerInfer~\cite{song2024powerinfer}, LLM in a flash~\cite{alizadeh2024llm} and Samoyeds~\cite{10.1145/3689031.3717455} exploit model sparsity to offload inactive weights. KV cache compression techniques~\cite{li2024snapkv,xiao2023efficient,zhang2023h2o,liu2023scissorhands,liu2024minicache,liu2024kivi,dong2024get} reduce memory footprint. \method extends the functionality of the hierarchical KV cache to support an arbitrary preemption pattern.

%% file: text/9_conclusion.tex
\section{Conclusion} \label{sec:8_conclusion}
We presented \method, an optimized LLM serving system that significantly enhances LLM text streaming performance through buffer-aware scheduling and hierarchical KV cache management. By dynamically aligning token generation rates with user consumption patterns and proactively managing GPU memory, \method achieved up to 82.5\% higher effective throughput and 80.2\% shorter time-to-first-token (TTFT). Extensive experiments demonstrated that \method outperforms state-of-the-art systems across diverse workloads and hardware configurations while sustaining smooth streaming quality. These results establish \method as a robust and efficient solution for real-time LLM applications.